\documentclass[10pt,twocolumn,letterpaper]{article}

\usepackage[pagenumbers]{iccv} %

\usepackage[utf8]{inputenc} %
\usepackage{url}            %
\usepackage{booktabs}       %
\usepackage{amsfonts}       %
\usepackage{amsmath}
\usepackage{amssymb}
\usepackage{rotating}
\usepackage{tikz}
\usepackage{pifont}
\usepackage{bbm}
\usepackage{nicefrac}       %
\usepackage{microtype}      %
\usepackage{xcolor}         %
\usepackage{graphicx}
\usepackage{multirow}
\usepackage{tabularx}
\usepackage{caption}
\usepackage{subcaption}
\usepackage{framed}

\usepackage{fancyvrb}
\usepackage{tcolorbox}
\usepackage{lipsum}

\newcommand{\cmark}{\ding{51}}%
\newcommand{\xmark}{\ding{55}}%

\newenvironment{smalltabularx}[1]{%
  \footnotesize
  \setlength{\tabcolsep}{4pt}
  
  \tabularx{\columnwidth}{#1}
}{%
  \endtabularx
}

\definecolor{pastelPink}{rgb}{1.0, 0.8, 0.87}
\definecolor{pastelGreen}{rgb}{0.79, 1.0, 0.79}

\tcbset{
  myboxstyle/.style={
    fonttitle=\scriptsize,
    fontupper=\scriptsize,
    coltitle=black,  %
    boxrule=0.5mm,
    width=\textwidth,
    left=1mm,
    right=1mm,
    top=1mm,
    bottom=1mm,
    colback=gray!10,  %
    colframe=gray!10  %
  },
  pastelPink/.style={
    colback=pastelPink,
    colframe=pastelPink,
    coltitle=black,
    fonttitle=\scriptsize,
    fontupper=\scriptsize,
    boxrule=0.5mm,
    width=\textwidth,
    left=1mm,
    right=1mm,
    top=1mm,
    bottom=1mm,
  },
  pastelGreen/.style={
    colback=pastelGreen,
    colframe=pastelGreen,
    coltitle=black,
    fonttitle=\scriptsize,
    fontupper=\scriptsize,
    boxrule=0.5mm,
    width=\textwidth,
    left=1mm,
    right=1mm,
    top=1mm,
    bottom=1mm,
  }
}

\newif\ifshowchanges
\showchangesfalse        %

\ifshowchanges
  \newcommand{\chg}[1]{\textcolor{blue}{#1}}
\else
  \newcommand{\chg}[1]{#1}
\fi

\definecolor{iccvblue}{rgb}{0.21,0.49,0.74}
\usepackage[pagebackref,breaklinks,colorlinks,allcolors=iccvblue]{hyperref}

\def\paperID{10361} %
\def\confName{ICCV}
\def\confYear{2025}

\title{Large-scale Pre-training for Grounded Video Caption Generation}

\author{Evangelos Kazakos$^{1}$, 
        Cordelia Schmid$^{2}$, 
        Josef Sivic$^{1}$ \\
        \\
        $^{1}$Czech Institute of Informatics, Robotics and Cybernetics at the Czech Technical University in Prague \\
        $^{2}$Inria, \'Ecole normale sup\'erieure, CNRS, PSL Research University\\\\
        {\small \hypersetup{urlcolor=magenta} \url{https://ekazakos.github.io/grounded_video_caption_generation/}}
        }

\begin{document}

\twocolumn[{
\maketitle
\vspace*{-8mm}
\centering
  \includegraphics[width=\textwidth]{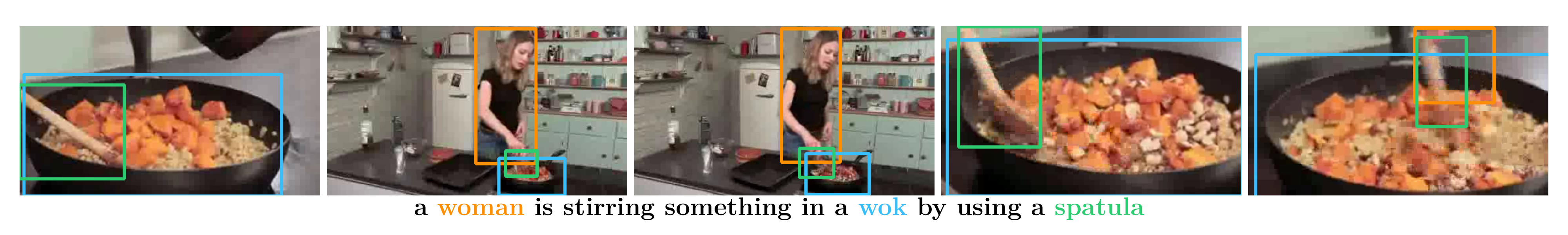}  
\vspace*{-7mm}
\captionof{figure}
{Output of our {\bf GROunded Video caption gEneration (GROVE)} model on an instructional video. The model outputs a video-level caption (bottom) with key noun phrases in the caption coloured and localised (grounded) in the video by temporally consistent bounding boxes (top). Note how the objects are consistently annotated (with the same color) despite changes in scale and viewpoint and how the person is marked as occluded (orange box not present) in frames 1 and 4 when the person (or their hand) is not visible.      
}
\label{fig:teaser}
\vspace{0.55cm}
}]

\maketitle
\linespread{0.96}
\begin{abstract}
\vspace{-9mm}

We propose a novel approach for captioning and object grounding in video, where the objects in the caption are grounded in the video via temporally dense bounding boxes.  We introduce the following contributions.
First, we present a large-scale automatic annotation method that aggregates frame-level captions grounded with bounding boxes into temporally dense and consistent annotations. We apply this approach on the HowTo100M dataset to construct a large-scale pre-training dataset, named HowToGround1M. We also introduce a Grounded Video Caption Generation model, dubbed GROVE, and pre-train the model on HowToGround1M. 
Second, we introduce iGround--a dataset of 3513 videos with manually annotated captions and dense spatio-temporally grounded bounding boxes. 
This allows us to measure progress on this challenging problem, as well as to fine-tune our model on this small-scale but high-quality data.
Third, we demonstrate that our approach achieves state-of-the-art results on the proposed iGround dataset, as well as on the VidSTG, ActivityNet-Entities, GroundingYouTube, and YouCook-Interactions datasets.
Our ablations demonstrate the importance of pre-training on our automatically annotated HowToGround1M dataset followed by fine-tuning on the manually annotated iGround dataset and validate the key technical contributions of our model. The dataset and code are available at \url{https://ekazakos.github.io/grounded_video_caption_generation/}.

\end{abstract}

\section{Introduction}
\label{sec:intro}

We seek to generate grounded captions for a given input video. As illustrated in Figure~\ref{fig:teaser}, the task is challenging 
as it entails both (i) generating the natural language caption for the video and (ii) predicting temporally dense bounding boxes for multiple noun phrases from the caption. Compared to grounded captioning in still images~\cite{zhang2023llavagrounding,peng2023kosmos2,zhao2023bubogpt,chen2023shikra,ma2024groma}, the task in the video domain has the additional difficulty that objects might disappear in some frames because of occlusions and we need to produce temporally dense and consistent bounding boxes across the frames of the input video.  
This problem is important as spatio-temporal grounding of individual objects with natural language descriptions on a large scale is one of the key steps to advance areas such as human-robot interaction and embodied perception~\citep{li2022estimating,mccarthy2024towards,Radosavovic2022,sermanet2018time,zorina2021learning}.  

Despite the advances on the grounded video caption generation problem~\cite{Ma2020Learning,Zanfir2016Spatio,zhou2019grounded} one of the key limiting factors hindering further progress is the lack of suitable large-scale videos datasets with captions densely grounded with multiple spatio-temporal boxes in the video.   
Existing datasets are restricted to localising a single spatio-temporal tube for each short textual description~\cite{hcstvg,vidstg,tanCOMMA2021,chen2024whatwhenwhere}, have limited temporal consistency as they provide bounding boxes for only a few sparsely sampled frames per video~\cite{zhou2019grounded,Ji_2020_CVPR,Grauman_2022_CVPR}, or are limited to a specific domain such as egocentric videos~\cite{VISOR2022,Liu_2022_CVPR,Grauman_2022_CVPR}.

In this work, we address this key limitation by the following three contributions. 
\textbf{First}, to address the issue of limited training data, we introduce a large-scale automatic annotation method leveraging an existing model for grounded still-image captioning together with an LLM to summarize frame-level captions into video-level captions. The LLM is also tasked to perform \emph{temporally consistent bounding box annotation}, associating frame-level phrases that correspond to objects with video-level phrases, resulting in a video-level caption grounded with multiple object tubes with consistent natural language labels.  We apply this approach to videos from the HowTo100M~\cite{miech19howto100m} dataset, which results in a new large-scale pre-training dataset, namely HowToGround1M, of 1M videos for this problem.  
This automatic annotation method is coupled with a proposed GROunded Video Caption gEneration model, called GROVE. The key technical contributions of this model include: (i) spatio-temporal adapters, which enable efficient modeling of spatio-temporal information in video; (ii) a bounding box decoder that outputs temporally coherent bounding boxes in video and (iii) a temporal objectness head that explicitly models objects that temporary leave the frame or are occluded. We pre-train the GROVE model on the proposed large-scale automatically annotated HowToGround1M dataset. 
\textbf{Second}, we introduce a new manually annotated dataset for the grounded caption generation task, which we name iGround. The dataset contains 3513 videos and more than 230,000 annotated object bounding boxes. We split this dataset into train/val/test sets (2013/500/1000, respectively). This allows us to measure progress on this challenging problem, as well as to fine-tune our model on small-scale but high-quality data. 
\textbf{Third}, our results demonstrate that our GROVE model achieves state-of-the-art performance on the proposed iGround dataset, as well as on the VidSTG~\cite{vidstg}, ActivityNet-Entities~\cite{zhou2019grounded}, GroundingYouTube~\cite{Chen_2024_CVPR}, and YouCook-Interactions~\cite{tanCOMMA2021} datasets.
We perform extensive ablations that demonstrate the importance of pre-training using our automatically annotated HowToGround1M dataset followed by fine-tuning on the manually annotated iGround dataset and validate the key technical contributions of our model. \textit{For additional details, see the appendix}.

\section{Related Work}
\label{sec:related}

\begin{table}[t]
\centering
\resizebox{\columnwidth}{!}{%
\begin{tabular}{lccccc}
    \toprule
    Dataset & \multicolumn{1}{p{1.2cm}}{\centering Annot. \\ type} &  \multicolumn{1}{p{1.2cm}}{\centering Multiple \\ frames} & \multicolumn{1}{p{1.8cm}}{\centering Multi-object \\ grounding} & \multicolumn{1}{p{1.2cm}}{\centering Num. \\ videos} & \multicolumn{1}{p{1.2cm}}{\centering Num. \\ instances} \\ 
    \midrule
    VidSTG~\citep{vidstg} & Manual & \checkmark & \xmark & 36.2K & \underline{9.9M} \\ 
    HC-STVG~\citep{hcstvg} & Manual & \checkmark & \xmark & 10.1K & 1.5M \\ 
    ActivityNet-Entities~\citep{zhou2019grounded} & Manual & \xmark & \checkmark & \underline{37.4K} & 93.6K \\
    {\bf HowToGround1M (Ours)} & Automatic & \textbf{\checkmark} & \textbf{\checkmark} & \textbf{1M} & \textbf{80.1M} \\ 
    {\bf iGround (Ours)} & Manual & \textbf{\checkmark} & \textbf{\checkmark} & 2K & 236.9K\\
    \bottomrule
\end{tabular}
}
\vspace*{-3mm}
\caption{
Comparison of our two datasets iGround and HowToGround1M with state-of-the-art video grounding datasets.} 
\label{tab:dataset_comparison}
\vspace*{-4mm}
\end{table}

\begin{figure*}[htpb]
  \includegraphics[width=\textwidth]{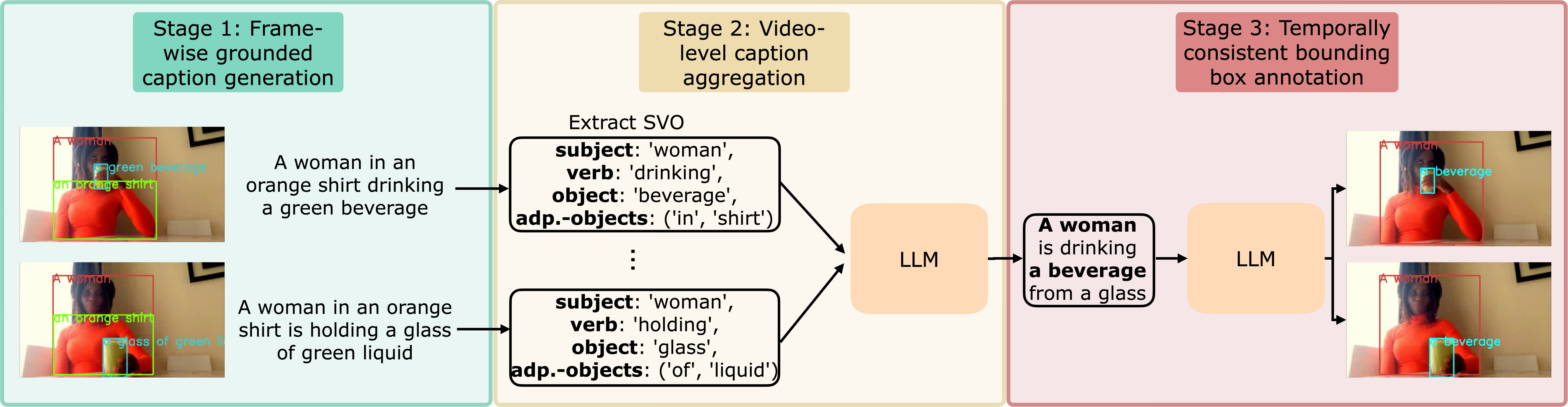}
  \vspace*{-7mm}
  \caption{{\bf A method for automatic annotation of spatio-temporally grounded captions.} In the first stage (left), we apply a still-image grounded caption generation model on individual video frames producing temporally inconsistent outputs. In the second stage (middle), the captions from individual frames are aggregated into a single video-level caption describing the most salient actions/objects in the video. Third (right), individual frame-level phrases and bounding boxes are associated over time into a temporally consistent and dense labelling of object bounding boxes over the video. }
  \label{fig:pseudolabelling}
  \vspace*{-5mm}
\end{figure*}

\textbf{Image-based grounded data generation}. Recent efforts in multi-modal learning have focused on grounding text to images and comprehending referring expressions~\citep{zhang2023llavagrounding,peng2023kosmos2,zhao2023bubogpt,chen2023shikra,ma2024groma,hanoona2023GLaMM}. Given the scale of these models, large-scale training data are essential, yet manual annotation is prohibitively expensive. To address this, methods typically leverage pre-trained models for automatic annotation.
Several approaches rely on LLM-driven data generation~\cite{zhang2023llavagrounding,chen2023shikra}, where GPT-4~\cite{openai2024gpt4} is used to create grounded dialogue datasets. Some methods pair instruction-tuned captions with bounding boxes~\citep{zhang2023llavagrounding}, while others generate QA pairs from existing bounding boxes and captions~\citep{chen2023shikra}. Other methods leverage NLP techniques, such as extracting noun chunks via Part-of-Speech tagging and aligning them with bounding boxes using a pre-trained grounding model~\citep{peng2023kosmos2}. More complex multi-stage annotation pipelines integrate multiple pretrained models to generate large-scale pseudolabeled datasets~\citep{hanoona2023GLaMM}. We build on this line of work but extend it to the video domain.

\noindent \textbf{Datasets for spatio-temporal grounding in video}. 
The existing datasets~\citep{hcstvg,vidstg,huang2018finding,tanCOMMA2021,chen2024whatwhenwhere,zhou2019grounded} for spatio-temporal video grounding are relatively small scale as they rely on manual annotation, which is tedious and time consuming. Typically, the existing datasets also focus on localizing a single spatio-temporal tube for each short textual description ~\citep{hcstvg,vidstg,tanCOMMA2021,chen2024whatwhenwhere}, which can be a limiting factor in instructional videos where multiple objects are often manipulated. 
In contrast, both our automatically annotated HowToGround1M dataset as well as our manually-labelled iGround dataset contain \textit{multiple} spatio-temporal bounding boxes grounding multiple objects described in the caption.
We compare our HowToGround1M dataset as well as our iGround dataset with exisiting video grounding  datasets in Table~\ref{tab:dataset_comparison}. ActivityNet-Entities~\cite{zhou2019grounded} is closest to ours in that it provides grounded captions -- multiple objects per frame for  annotated noun phrases in the caption. Nevertheless, the authors annotate a single frame per object in the video segment while both our automatically annotated HowToGround1M as well as our manually annotated iGround dataset have densely annotated frames per video segment. Moreover, HowToGround1M has the largest number of videos across all datasets and the largest number of annotated instances.
More statistics and analysis of our proposed datasets is provided in appendix~\ref{sec:dataset_stats}.%

\noindent \textbf{Spatio-temporal grounding in video}. Spatio-temporal grounding~\cite{hcstvg,vidstg,yang2022tubedetr,tanCOMMA2021,Lin_2023_CVPR,chen2024whatwhenwhere,Su_2021_ICCV,ijcai2020p149,Yang_Neurips_2022,Gu_2024_CVPR,Wasim_2024_CVPR} aims to predict a single spatio-temporal tube enclosing an event described in a natural language query. Early approaches relied on object detection features~\cite{vidstg,hcstvg}. \cite{vidstg} introduced a spatio-temporal graph encoder, while \cite{hcstvg} and \cite{yang2022tubedetr} adapted transformers for multi-modal grounding. Some works employed contrastive learning on large-scale instructional videos, leveraging weak supervision from HowTo100M~\cite{tanCOMMA2021,chen2024whatwhenwhere}. Others explored architectural innovations, such as two-stream models for appearance and motion~\cite{Lin_2023_CVPR} or context-guided decoding for object-centric grounding~\cite{Gu_2024_CVPR}. However, these methods lack a text generation component, as they assume the query is given, and they cannot generate multiple spatio-temporal tubes for multiple objects. Our approach addresses this issue with the  GROVE model, a novel automatic annotation method, and a manually-labeled dataset for fine-tuning and evaluation. More pertinent to our work is~\cite{zhou2019grounded}, which separately performs captioning and grounding. In contrast, our GROVE model jointly (i) captions and (ii)~grounds noun phrases with temporally dense bounding boxes while handling occlusions. \chg{Concurrent to our work is
\cite{Munasinghe_2025_CVPR}, which targets mask-level video grounding and trains on semi-automatic annotations from existing benchmarks. In contrast, we focus on bounding boxes, pre-train with large-scale fully automatic annotations and fine-tune on human-annotated data. 
We outperform both~\cite{Munasinghe_2025_CVPR} and~\cite{zhou2019grounded} .}

\section{Large-scale generation of grounded captions}

\label{sec:pseudolabeling}

\begin{figure*}[t]
 \centering
  \includegraphics[width=\textwidth]{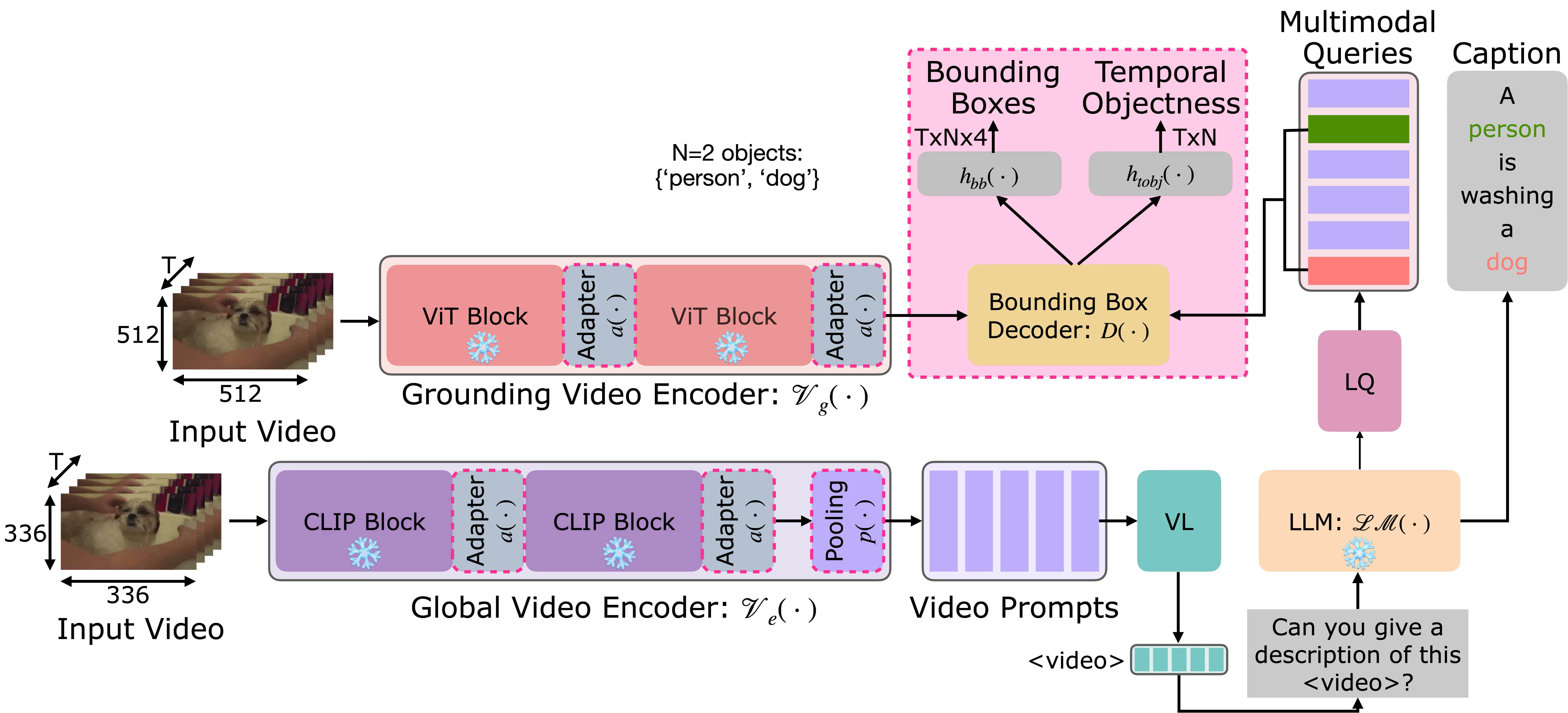}
  \vspace*{-5mm}
  \caption{{\bf An overview of our GROVE model for grounded video caption generation.} 
  Dashed red rectangles outline the key technical contributions enabling grounded caption generation in video and include: (i) spatio-temporal adapters; (ii) the bounding box decoder and (iii) the temporal objectness head. \chg{VL: vision-to-language projection, LQ: language-to-query projection.}}
  \label{fig:groc_model}
  \vspace*{-3mm}
\end{figure*}

In this section, we first introduce our automatic annotation method for generating a large-scale dataset for grounded video caption generation (Sec.~\ref{subsec:automatic_annotation_method}). We then describe HowToGround1M (Sec.~\ref{subsec:pretrain_datset}), our dataset created by applying this method to videos from HowTo100M~\cite{miech19howto100m,shvetsova2023howtocaption}.

\subsection{Automatic annotation method}
\label{subsec:automatic_annotation_method}

We describe our method for generating an automatically annotated dataset for grounded video caption generation. Given an unlabelled dataset of videos depicting humans interacting with objects and/or other humans, the goal of the method is to generate both video-level captions describing what is happening in the video \textit{and} temporally dense and consistent bounding boxes grounded to the noun phrases from the caption that describe the main objects in the video. We leverage foundation LMMs and LLMs as they have been pretrained on large-scale datasets and provide a rich source of information. Our method consists of three steps, as shown in Figure~\ref{fig:pseudolabelling}: i) frame-wise grounded caption generation, ii) video-level caption aggregation and iii) temporally consistent annotation of objects. We describe these steps next.  

\noindent\textbf{Stage 1: Frame-wise grounded caption generation.} We begin by generating grounded captions for individual video frames, producing text descriptions alongside bounding boxes for noun phrases referring to objects. As no video models exist for this task, we use image-based models in a frame-by-frame manner. We adopt GLaMM~\citep{hanoona2023GLaMM}, which excels in image-based grounded captioning. Since GLaMM outputs segmentation masks, we convert them to bounding boxes. An example output is shown in Figure~\ref{fig:pseudolabelling} (left). 
In the experiments, we also compare an alternative approach that combines frame-level captioning~\cite{wang2022git} with open-vocabulary object detection~\cite{minderer2024scalingopenvocabularyobjectdetection} to obtain candidate captions and bounding boxes in individual frames.

\noindent\textbf{Stage 2: Video-level caption aggregation.} 
The previous stage provides outputs for individual frames, which are, however, not temporally consistent. We address this issue in Stage 2.
From the frame-level captions, we generate a video-level caption that highlights the most salient actions and objects while ensuring consistency in noun phrase annotations across frames. We achieve this by prompting Llama-2~\citep{touvron2023llama}, as described next.
Since the grounded captioning model produces lengthy captions with extraneous details, we first extract Subject-Verb-Object (SVO) triplets and relevant adpositional phrases using Part-of-Speech (POS) tagging. These structured triplets serve as input to the LLM, encapsulating subjects, actions, and relevant objects. To enhance accuracy, we perform in-context learning by providing example pairs of frame-level SVO triplets and their corresponding video-level captions (see Figure~\ref{fig:stage2} for details). Given a new set of SVO triplets, the LLM generates a video-level caption and tags noun phrases corresponding to objects of interest within \texttt{<p></p>} tags. Figure~\ref{fig:pseudolabelling} (middle) illustrates this process.
In experiments, we compare the above approach with providing the LLM with full captions from Stage 1 instead of extracting Subject-Verb-Object triplets from the caption to assess the impact of additional context.

\noindent\textbf{Stage 3: Temporally consistent bounding box annotation.} While Stage 2 ensures a consistent video-level caption, noun phrase annotations remain inconsistent across frames due to the use of an image-based model. To resolve this, we introduce \textit{temporal labeling of objects}, ensuring that bounding boxes corresponding to the same object are consistently labeled throughout the video using the video-level noun phrases. These consistently labeled bounding boxes form video object tracks.
We formulate this as a text classification task and prompt the LLM with in-context learning. The input consists of frame-level noun phrases to be classified and the video-level noun phrases serving as class labels. Figure~\ref{fig:stage3} provides details on the prompt. Figure~\ref{fig:pseudolabelling} (right) illustrates this process.
An alternative approach is to use a visual tracker, such as~\cite{karaev24cotracker3}, for associating bounding boxes over time, which we compare against in Sec.~\ref{subsec:res}. 

Upon completing all three stages, we obtain videos with automatically generated captions, grounded bounding boxes, and temporally coherent labels aligned with the noun phrases in the caption. Additional details are provided in appendix~\ref{sec:automatic_annotation_method_details}.

\subsection{The HowToGround1M dataset}
\label{subsec:pretrain_datset}

We apply our automatic annotation method to Internet instructional videos from the HowTo100M dataset~\cite{miech19howto100m}. We choose HowTo100M due to its diversity of actions, scenes, objects and lighting conditions. We apply our approach to clips obtained using time stamps from the HowToCaption~\cite{shvetsova2023howtocaption} version of the dataset. These clips contain meaningful events in the video identified based on LLM analysis of the associated narrations. %
In detail, we randomly sample 1M video clips from HowTo100M videos using start/end timestamps from HowToCaption. \chg{We ensured there is no video ID overlap with any of the downstream datasets, including iGround.
}
The videos from HowTo100M have variable frame rates usually ranging in 25-30 fps, and we process the videos at 5fps. The majority of the clips are 8 seconds long with a spatial resolution of 455$\times$256 pixels. We run our automatic annotation method on this set of data to obtain our HowToGround1M pre-training dataset. The resulting dataset contains 1M videos, with 1M captions containing 3.2M noun phrases. The captions contain 18.6k unique terms and 142k unique noun phrases. Overall, the dataset contains 43.6M annotated frames with 80.1M bounding boxes. This dataset is used to pre-train the GROVE model, described next.

\section{The GROVE Model}
\label{sec:groc_model}

We introduce a GROunded Video caption gEneration model, called GROVE, see Figure~\ref{fig:groc_model}.
The input to the model is a video clip with $T$ frames (left) and the output is a natural language caption (right) together with $N$ spatio-temporal bounding boxes localizing the individual noun phrases in the video together with a objectness score indicating the presence / absence of the object in a specific frame (top).
We build on the GLaMM~\citep{hanoona2023GLaMM} model, which is a state-of-the-art method for still image-based grounded caption generation, and extend it to the video domain. 
The key technical components enabling grounded caption generation in videos are (shown in dashed red rectangles in Figure~\ref{fig:groc_model}): i) the \textbf{spatio-temporal adapters with pooling} which enable modelling temporal information efficiently; ii) the \textbf{bounding box decoder} which allows re-using large-scale pretrained decoder weights~\cite{hanoona2023GLaMM}; and iii) the \textbf{temporal objectness head} for modelling objects that  temporary leave the frame or are occluded. Details are given below. We assess the importance of these components in Section~\ref{sec:experiments}. Additional details of the GROVE model and the training procedure are in appendix~\ref{sec:GROC_details}.

\vspace*{1mm}
\noindent\textbf{Spatio-temporal adapters and pooling}. 
The visual information is encoded using the Global Video Encoder $\mathcal{V}_e(\cdot)$, which represents the video globally for captioning, and the Grounding Video Encoder $\mathcal{V}_g(\cdot)$, which represents the fine-grained details for grounding. 
We build these encoders %
by adapting the respective pre-trained image encoders~\cite{hanoona2023GLaMM}. We achieve this by interleaving spatio-temporal adapter layers (denoted as $a()$) between the image-based encoder layers. To stabilise training, we add residual connections and introduce a learnable parameter that is multiplied by the adapter's output and starts from 0 at the beginning of training~\citep{Flamingo_NEURIPS2022}. By doing so, at the beginning of training the adapter's output is effectively cancelled out and the network observes only the original encoder's output. As training progresses, the learnable parameter is tuned and the network automatically adjusts the contribution of the adapter based on the gradients of the loss. 
In detail, the adapter layer performs $ a(o) = o + tanh(\alpha)\times f(o)$, 
where $o$ is the output of the preceding encoder layer, $\alpha$ is the tunable parameter that is initialised to 0 and passes through a $tanh$ activation and $f(\cdot)$ is the adapter layer.
As feeding the full video tokens $o_e$ to the LLM is computationally prohibitive, we introduce a spatio-temporal pooling function after the output of $\mathcal{V}_e(\cdot)$, \textit{i.e.}, $o_p=p(o_e)$.

\noindent\textbf{Bounding box decoder and prediction head}. We adapt the pre-trained mask decoder~\citep{kirillov2023segment,hanoona2023GLaMM} for bounding box decoding. We focus on bounding boxes (rather than full pixel-level masks) as they are easier and cheaper to manually annotate yet provide good localization accuracy for compact objects, which are the main focus of this work.  
We transform the mask decoder to a bounding box decoder by using the embedded detection tokens as queries, and the visual features of the Grounding Video Encoder as keys/values, resulting in an output that has same length as the detection tokens, allowing us to predict a bounding box for each detection token that corresponds to a noun phrase in the caption. Importantly, while $\mathcal{V}_g(\cdot)$ performs video processing, we apply the cross-attention in a frame-wise fashion to predict objects at each frame of the input video.
We employ a bounding box prediction head on the output of the bounding box decoder, $o_d$. It is an MLP that predicts bounding box coordinates for the embedded detection tokens at each frame: $p_{bb}=h_{bb}(o_d)$, where $p_{bb} \in \mathbb{R}^{T\times N_d\times 4}$ are the bounding box predictions and $h_{bb}(\cdot)$ is the bounding box head.

\noindent\textbf{Temporal objectness head}. As discussed previously, one major challenge for videos is that objects might disappear and reappear in different frames of the video. To address this, we introduce a \textit{temporal objectness head}. Different than objectness predictions in image-based object detection, the purpose of this head is to predict whether an object is visible or not at a given frame of a video: $p_{tobj}=h_{tobj}(o_d)$, where $p_{tobj} \in \mathbb{R}^{T\times N_d\times 1}$ are the temporal objectness scores and $h_{tobj}(\cdot)$ is the temporal objectness head. During inference, we threshold $p_{tobj}$ and for each frame we select only the bounding boxes for which the temporal objectness scores pass the threshold.

 \section{Manually annotated iGround dataset}
\label{sec:howtoground_manual_dataset}

For the iGround dataset, we select \chg{3513} clips from the HowTo100M dataset~\citep{miech19howto100m} by sampling `interesting' videos that typically include dynamic events or actions that are clear and distinguishable. In those events/actions, people usually interact with objects. We make sure that the 3513 clips do not overlap with those in the HowToGround1M dataset described in section~\ref{sec:pseudolabeling}. \chg{Moreover, we ensure that there is no video ID overlap with any of the downstream datasets.
We split the set into a training set (2013 clips), a validation set (500 clips), and a test set (1000 clips), ensuring no overlap in video IDs between them.}
The video annotation itself consists of 3 steps. The first step entails watching the video and providing a natural language description of what is happening in the video and the objects that are being manipulated. Note, that we are interested in the \textit{active objects}, \textit{i.e.} objects that humans interact with, rather than densely describing all objects in the scene. In the second step, bounding boxes are annotated for all visible instances of humans/objects mentioned in the caption that has been provided in the previous step. 
Finally, each bounding box is annotated with a short phrase or word that should match exactly a short phrase/word from the caption. More information on the annotation procedure, including annotation guidelines for the raters and mechanisms to ensure the consistency and overall quality of the raters' annotations, can be found in appendix~\ref{sec:annotation_protocol}.

\section{Experiments}
\label{sec:experiments}
In this section we introduce evaluation datasets and metrics,  compare the proposed approach with the state of the art on three benchmark datasets, analyze the effect of the pre-training dataset size and ablate the key components of the proposed model and automatic annotation procedure.

\subsection{Datasets and evaluation metrics}
\label{sec:datasets}
We evaluate the proposed approach on three datasets, the newly introduced grounded video caption generation dataset iGround as well as the established VidSTG~\cite{vidstg} and ActivityNets-Entities~\cite{zhou2019grounded} video grounding benchmarks.

\noindent\textbf{iGround}. We build on the metrics for grounding captions in still images~\cite{hanoona2023GLaMM} and adapt them to our task in videos. These include METEOR~\citep{banerjee-lavie-2005-meteor} and CIDEr~\citep{Vedantam_2015_CVPR} for the quality of the captions, AP50 for the grounding accuracy, and recall~\citep{hanoona2023GLaMM} that combines (i) IoU between ground truth (GT) and predicted bounding boxes as well as (ii) the similarity of embeddings of GT and predicted noun phrases that correspond to bounding boxes. The aim of the recall metric is to assess the rate of positive predictions. A prediction is considered positive if both the bounding box IoU and the noun phrase similarity are above a certain threshold. We propose a video-level evaluation setting for AP50 and recall for the grounded video caption generation task where the metrics are calculated per video and averaged across videos.

\noindent\textbf{ActivityNet-Entities}. We follow \cite{zhou2019grounded} and report F1\textsubscript{all}, F1\textsubscript{all\_per\_sent}, F1\textsubscript{loc}, and F1\textsubscript{loc\_per\_sent}. In F1\textsubscript{all}, a region prediction is considered correct if the associated noun phrase is both correctly predicted (exact match) and correctly localised (IoU $>$ 0.5). F1\textsubscript{loc} considers only localisation accuracy ignoring errors in the generated noun phrases. In these metrics, accuracies are averaged across noun categories while in F1\textsubscript{all\_per\_sent} and F1\textsubscript{loc\_per\_sent} accuracies are averaged across sentences.

\noindent\textbf{VidSTG}. We follow~\cite{Su_2021_ICCV,yang2022tubedetr,Yang_Neurips_2022,zhou2023dense} and assume that the videos are temporally segmented to the events of interest (using the available start/end timestamps) and report m\textsubscript{sIoU}, defined as the average IoU across frames, between the predicted and ground truth bounding boxes for the target event.

\noindent\textbf{Implementation details}. All implementation details including architectural choices of the GROVE model as well as training and inference details can be found %
in appendix~\ref{sec:GROC_details}.

\begin{table}[t]
\centering
\begin{smalltabularx}{lXllll}
\toprule
& Method & METEOR & CIDER & AP50 & Recall \\
\midrule
\multirow{3}{*}{\rotatebox{90}{Center}} 
& a. GLaMM~\cite{hanoona2023GLaMM} & 11.9 & 29.9 & 20.8 & 19.3 \\
& b. GROVE - PT (Ours) & 14.3 & 50.6 & 27.0 & 22.5 \\
& c. GROVE - PT+FT (Ours) & \textbf{21.4} & \textbf{83.5} & \textbf{31.7} & \textbf{26.2} \\
\midrule\midrule
\multirow{4}{*}{\rotatebox{90}{All}} 
& d. Automatic annotation & 13.8 & 40.0 & 27.1 & 20.4 \\
& e. GROVE - PT (Ours) & 14.3 & 50.6 & 33.6 & 24.3 \\
& f. GROVE - FT (Ours) & 21.0 & 77.7 & 15.8 & 18.1 \\
& g. GROVE - PT+FT (Ours) & \textbf{21.4} & \textbf{83.5} & \textbf{40.0} & \textbf{28.7} \\
\bottomrule
\end{smalltabularx}
\vspace*{-3mm}
\caption{ {\bf Grounded video caption generation} on manually-annotated iGround test set. Pre-training on our new large-scale HowToGround1M dataset followed by finetuning on manually-annotated iGround training data (PT+FT)  clearly outperforms pre-training only (PT) and finetuning only (FT) as well as the GLaMM baseline~\citep{hanoona2023GLaMM} (a.) and directly applying automatic annotation (d.). We show center frame (``Center") and all frame (``All") evaluation.  }
\vspace*{-2mm}
\label{tab:results}
\end{table}

\begin{table}[t]
\begin{minipage}{0.49\columnwidth}
\centering
\begin{smalltabularx}{Xl}
\toprule
Method & m\textsubscript{sIoU} \\
\midrule
STVGBert~\cite{Su_2021_ICCV} & 47.3\\
TubeDETR~\cite{yang2022tubedetr} & 59.0\\
STCAT~\cite{Yang_Neurips_2022} & 61.7\\
DenseVOC~\cite{zhou2023dense}  & 61.9\\
GROVE FT (Ours) & 61.3\\
GROVE PT+FT (Ours) & \textbf{63.7}\\
\bottomrule
\end{smalltabularx}
\vspace*{-3mm}
\caption{State-of-the-art comparison of spatial grounding on the VidSTG~\cite{vidstg} test set (declarative sentences). All models
use ground truth temporal localization. Large-scale pretraining (PT+FT) results in an improvement over fine-tuning only (FT) for our model GROVE. }
\vspace*{-2mm}
\label{tab:vidstg}
\end{minipage}
\hfill
\begin{minipage}{0.49\columnwidth}
\centering
\begin{smalltabularx}{lXl}
\toprule
Method & FT & $\text{m}_{\text{sIoU}}$ \\
\midrule
PG-V-L (13B) \cite{pgvideollava} & \xmark & 35.1 \\
GLaMM~\cite{hanoona2023GLaMM} \\+ SAM2~\cite{ravi2025sam}  & \xmark & 38.6 \\
GROVE & \xmark  & \textbf{43.0} \\
\midrule
VideoGLaMM\cite{Munasinghe_2025_CVPR}  & \cmark & 39.7 \\
GROVE & \cmark & \textbf{55.5} \\
\bottomrule
\end{smalltabularx}
\vspace*{-3mm}
\caption{\chg{State-of-the-art comparison of spatial grounding on the VidSTG~\cite{vidstg} test set (interrogative sentences). All models use ground truth temporal localization. GROVE outperforms all competitors both in a pre-training only setting (\xmark) and when fine-tuned on VidSTG (\cmark).}}
\vspace*{-2mm}
\label{tab:vidstg_interrogative}
\end{minipage}
\end{table}

\begin{table}
\begin{minipage}{\columnwidth}
\centering
\begin{smalltabularx}{Xcccc}
\toprule
Method & F1\textsubscript{all} & F1\textsubscript{all\_per\_sent} & F1\textsubscript{loc} & F1\textsubscript{loc\_per\_sent}\\
\midrule
GVD~\cite{zhou2019grounded} & 07.10 & 17.30 & 23.80 & 59.20\\
GROVE FT (Ours) & 09.51 & 21.15 & 30.96 & 68.79 \\
GROVE PT+FT (Ours) & \textbf{13.39} & \textbf{24.08} & \textbf{45.04} & \textbf{77.29}\\
\bottomrule
\end{smalltabularx}
\vspace*{-3mm}
\caption{Results on the validation set of ActivityNet-Entities~\cite{zhou2019grounded}. Large-scale pretraining (PT+FT) results in an improvement over fine-tuning only (FT) for our model GROVE. }
\vspace*{-4mm}
\label{tab:anet}
\end{minipage}
\end{table}

\begin{figure*}[htbp]
 \centering
  \includegraphics[width=\textwidth]{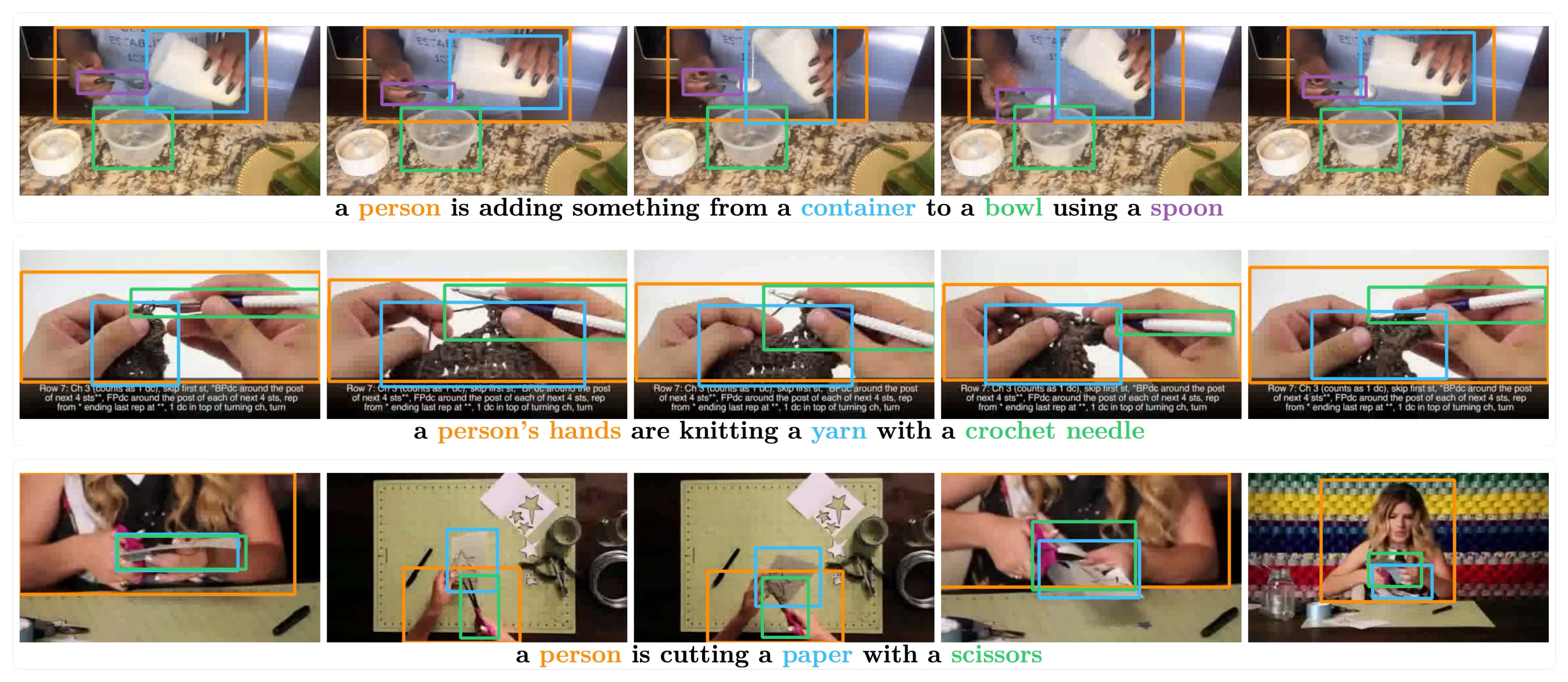}
  \vspace*{-8mm}
  \caption{{\bf Qualitative examples showing predictions of our GROVE model.} Please note that GROVE is able to: (i) produce video-level natural language captions describing the main action in the video; (ii) ground multiple objects; and (iii) produce spatio-temporally consistent bounding box predictions.
  Please note that the second row shows an example of model prediction that is partly incorrect as the blue box, while temporally consistent, does not depict a ``yarn". 
 {\bf More results including \chg{failure modes} are discussed in appendix~\ref{sec:qualitative}.}} %
  \label{fig:qual1}
  \vspace*{-2mm}
\end{figure*}

\begin{figure*}[t]
\centering
\includegraphics[width=\textwidth]{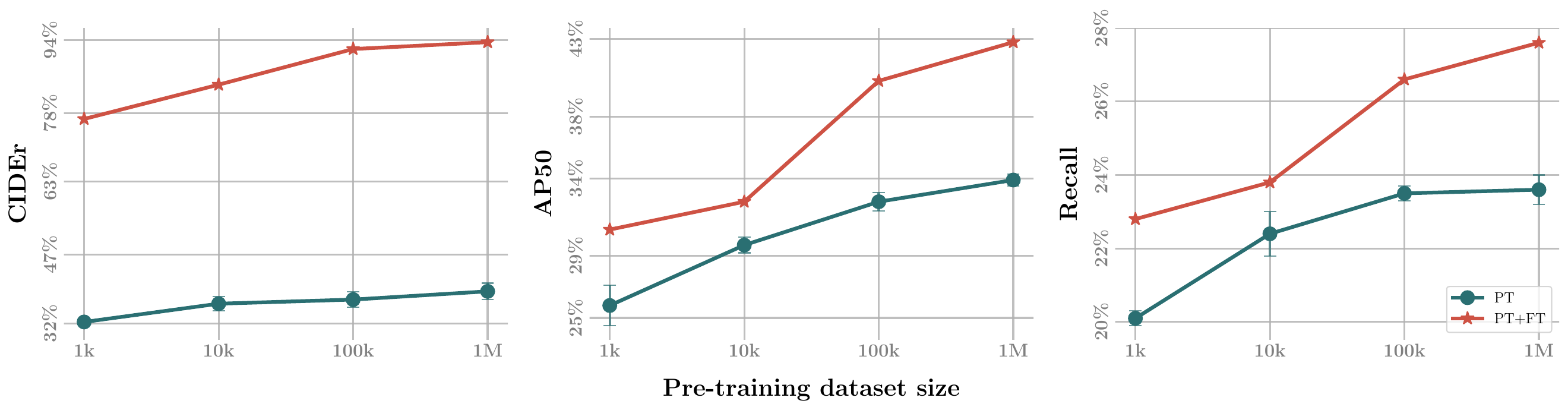}
  \vspace*{-8mm}
  \caption{Results after pre-training (PT) vs. after fine-tuning and pre-training (PT+FT) as a function of the pre-training dataset size. Results are reported on the iGround validation set.}
  \label{fig:scaling}
    \vspace*{-3mm}
\end{figure*}

\subsection{Comparison with the state of the art}
\label{subsec:res}

\textbf{iGround}. The results on our human-annotated iGround test set are shown in Table~\ref{tab:results}. We compare the results of the proposed GROVE model with our automatic annotation method (described in Section~\ref{sec:pseudolabeling}) and (still image)
GLaMM~\cite{hanoona2023GLaMM}. The automatic annotation method is a natural fit for a baseline for this task as it performs image-based grounded captioning followed by video-level aggregation without any training. Comparing with~\cite{hanoona2023GLaMM} aims to assess the benefits of our method in comparison to still-image grounding. Since \cite{hanoona2023GLaMM} runs per frame, it does not provide a video-level caption and noun phrases; these differ across frames (see Sec.~\ref{sec:pseudolabeling}). Thus, we use predictions of~\cite{hanoona2023GLaMM} for the center frame of each video. 
We call this the ``Center" set-up in Table~\ref{tab:results}. The ``All" set-up considers video-level caption and all frames of the video for bounding box localisation.
GROVE, pre-trained on our automatically annotated HowToGround1M dataset (b.), significantly outperforms the \cite{hanoona2023GLaMM} baseline (a.) with an improvement of 20 points in CIDEr and 6.2  points in AP50. This demonstrates that temporal context is crucial for this task. Results further improve with finetuning (c., PT+FT). 
In the all-frame evaluation, the GROVE model pre-trained on the automatically annotated HowToGround1M dataset (e.) provides significant performance improvements over directly obtaining predictions on the test set using the automatic annotation method (d.). This indicates that the GROVE model can correct (smooth out) some of the noise in the automatic annotations during the large-scale training. 
\chg{We obtain further major performance improvements by fine-tuning GROVE on our manually-annotated iGround dataset, showing that large-scale pre-training with automatic labels followed by fine-tuning on a small-scale but highly accurate dataset (PT+FT, c. and g.) is a good recipe for this task. Finally, the underperformance of GROVE that is directly fine-tuned on iGround without pre-training (FT, f.) provides further evidence of the importance of pre-training, particularly when the fine-tuning dataset is small.}

\noindent\textbf{ActivityNet-Entities and VidSTG}. We adapt the GROVE model for the spatio-temporal video grounding task on the  ActivityNet-Entities and VidSTG datasets. The details are in appendix~\ref{sec:GROC_details}.  
\chg{Tables~\ref{tab:vidstg},~\ref{tab:vidstg_interrogative}~and~\ref{tab:anet} show the comparison of GROVE with published results on VidSTG and ActivityNet-Entities.
On VidSTG declarative sentences (Table~\ref{tab:vidstg}), GROVE achieves the best performance despite not being designed for this task. On VidSTG interrogative sentences, (Table~\ref{tab:vidstg_interrogative}), GROVE outperforms
all previous work, both when it is fine-tuned on VidSTG (\cmark), but also when it is only pre-trained on HowToGround1M
pre-training set (\xmark).}
On ActivityNet-Entities (Table~\ref{tab:anet}), GROVE significantly outperforms the GVD baseline, demonstrating its effectiveness even with sparse annotations.  

\noindent\chg{\textbf{Other datasets}. In Table~\ref{tab:other_datasets}, we evaluate GROVE on YouCook-Interactions~\cite{tanCOMMA2021} and GroundingYouTube~\cite{Chen_2024_CVPR}, outperforming the previous SOTA by large margins.}

\noindent\textbf{Qualitative Results}. We show qualitative results of the GROVE model on several example videos in Figures~\ref{fig:teaser} and ~\ref{fig:qual1}. More qualitative results are in appendix~\ref{sec:qualitative}.

\subsection{Effects of pre-training dataset size}
\vspace{-2mm}

We study the scaling behaviour of the GROVE model by varying the size of the pre-training data.  Specifically, we pre-train GROVE on 1k, 10k, 100k and 1M videos sampled from our HowToGround1M dataset. The performance is measured on the iGround validation set, where we report CIDEr, AP50 and recall. The results are shown in Figure~\ref{fig:scaling}.
For each metric, we report the performance of the pre-trained model for different amounts of pre-training data, but also the performance of the fine-tuned model when initialised from the pre-trained models and fine-tuned on the manually annotated iGround training set. 
The results show that as we scale the pre-training dataset, both the pre-trained and the fine-tuned models continue to improve consistently across all metrics. This finding is important, as it verifies that obtaining automatic annotations at a large scale is beneficial for pre-training. It also signifies that fine-tuning using small-scale but high-quality data is most efficient when the model is pre-trained at the largest scale.

\noindent\textbf{Benefits of pre-training.} The benefits of large-scale pre-training on the HowToGround1M dataset are clearly demonstrated on three different tasks and datasets, as shown in Table~\ref{tab:results} (iGround), 
Table~\ref{tab:vidstg} (VidSTG)~and Table~\ref{tab:anet} (ActivityNet-Entities), where the model pre-trained on the HowToGround1M dataset and finetuned on each specific dataset reaches state-of-the-art results.

\begin{table}[t]
\begin{minipage}{\columnwidth}
\centering
\begin{smalltabularx}{XXllll}
\toprule
Unfreeze & AD & METEOR & CIDEr & AP50 & Recall \\
\midrule
\checkmark & \checkmark & \textbf{19.7} & \textbf{92.6} & \textbf{42.0} & \textbf{26.9} \\
\checkmark & \xmark &  \textbf{19.7} & 88.9 & 39.2 & 26.4 \\
\xmark & \checkmark & 19.2 & 82.2 & 36.8 & 25.9 \\
\bottomrule
\end{smalltabularx}
\vspace*{-3mm}
\caption{Ablation of spatio-temporal adapters (AD) and unfreezing the bounding box decoder and projection layers (unfreeze). We report results on the iGround validation set.}
\label{tab:ablations1}
\end{minipage}
\hfill

\begin{minipage}{\columnwidth}
\centering
\captionsetup{type=figure}
  \includegraphics[width=\columnwidth]{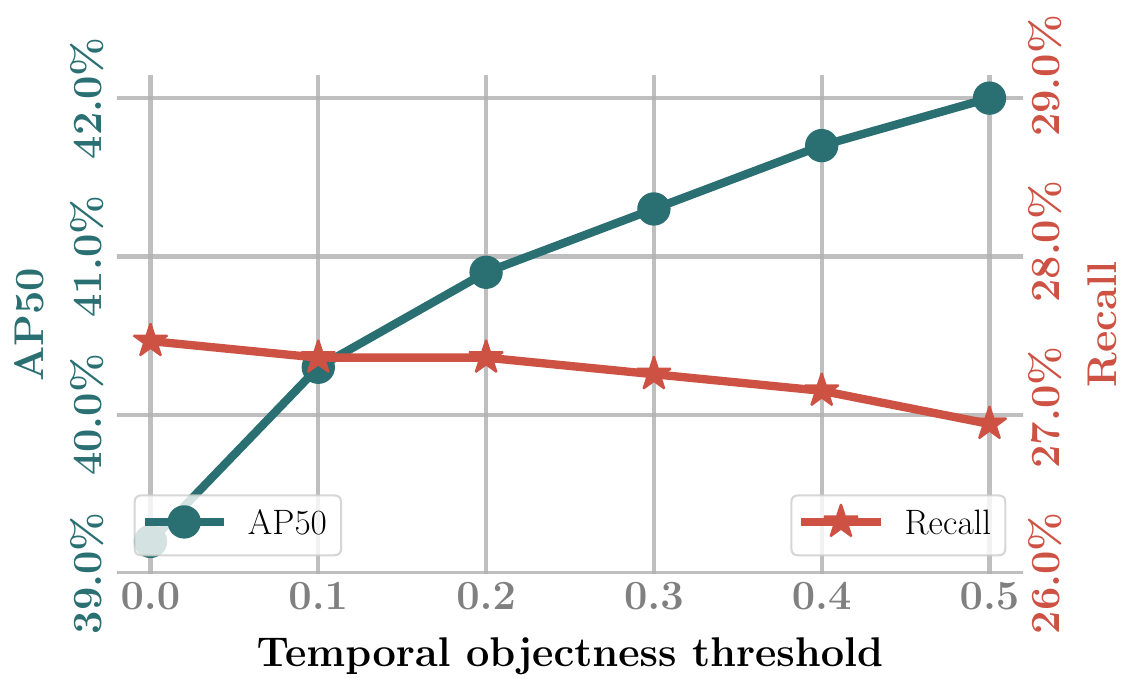}
  \vspace*{-7mm}
  \caption{{\bf Benefits of temporal objectness.} AP50 (left) and recall (right) of our model for different temporal objectness thresholds. Results are reported on the iGround validation set.}
  \label{fig:to_thres}
\end{minipage}

\begin{minipage}{\columnwidth}
\centering
\vspace*{4mm}
\begin{smalltabularx}{Xllllll}
\toprule
 Auto. annotation & METEOR & CIDER & AP50 & Recall & Average\\
\midrule
 a. {\bf Proposed} & 12.3 & 31.7 & 26.9 & 19.3 & \textbf{22.5}\\
 b. Alt. Stage 1 (F)& 12.4 & 36.8 & 20.1 & 14.5 & 20.9\\
 c. Alt. Stage 1 (V) & 16.8 & 23.1 & 14.6 & 14.2 & 17.2\\
 d. Alt. Stage 2 & 11.9 & 28.7 & 27.8 & 17.6 & 21.5\\
 e. Alt. Stage 3 & 12.3 & 31.7 & 23.2 & 16.0 & 20.8\\
\bottomrule
\end{smalltabularx}
\vspace*{-3mm}
\caption{Comparison of our automatic annotation approach (row a.) vs. several alternatives (rows b.-e.) on the iGround validation set. See text for details. 
}
\label{tab:results_alternative_stage1}
\end{minipage}
\vspace*{-4mm}
\end{table}

\subsection{Ablation analysis}
Here we provide the ablation of the key components of the GROVE model as well as evaluate variants of the automatic annotation method.

\noindent\textbf{Model ablations}. We ablate the key components of our GROVE model. We report results on the iGround validation set after pre-training with fine-tuning, which achieves the best results. 
In Table~\ref{tab:ablations1}, we ablate the spatio-temporal adapters and the training (unfreezing) of the bounding box decoder and projection layers (VL and LQ in Fig.~\ref{fig:groc_model}). To mitigate overfitting, we keep the visual backbones and LLM frozen, while training the embedding and output layers of the LLM to accommodate the modified vocabulary.

Next, we assess the importance of the temporal objectness head. Figure~\ref{fig:to_thres} shows the performance of the GROVE model in AP50 and recall when varying the threshold for the temporal objectness from 0.0 (completely removing the temporal objectness head) to 0.5. This threshold regulates the sensitivity of the model to detecting whether an object leaves the frame (or is occluded). These results demonstrate that by increasing the temporal objectness threshold we obtain substantial benefits in AP50 while sacrificing only a negligible amount of recall. This demonstrates the importance of modelling objects that temporally leave the frame. %

\noindent\textbf{Ablations of automatic annotation}.
\chg{We ablate each stage of our automatic annotation method and show results in Table~\ref{tab:results_alternative_stage1}. In ``b. Alt. Stage 1 (F)'', we replace our Stage 1 with the GIT~\citep{wang2022git} image captioner + Llama-3~\citep{dubey2024llama3herdmodels} noun phrase extractor + OWLv2~\citep{minderer2024scalingopenvocabularyobjectdetection} object detector, which yields slightly crisper captions but weaker grounding because the captioner and detector are not trained jointly; a video-level variant (VideoLlama-3~\cite{zhang2025videollama3} instead of GIT), row ``c. Alt. Stage 1 (V)'', shows the same pattern. Feeding full captions in Stage 2 (row ``d. Alt. Stage 2'') instead of using subject-verb-object triplets (row ``a. Proposed'') raises AP50: full-captions yield fewer predictions as the LLM trims its output to the most salient objects, reducing recall but improving precision--hence the slightly higher AP50. Adding CoTracker3 in Stage 3 (row ``e. Alt. Stage 3'') further degrades grounding due to tracker drift during viewpoint changes. Our proposed automatic annotation method (row ``a. Proposed'') scores the best on average; a more detailed analysis is in appendix~\ref{sec:quantitative}.}

\section{Conclusion}
\vspace{-2mm}

We introduced a large-scale automatic annotation method to generate densely grounded captions in video, which we used to construct the large-scale HowToGround1M video dataset. We developed GROVE, a new model for grounded video captioning, which we pre-trained on this large-scale data. To enable rigorous evaluation, we introduced iGround, a dataset with high-quality manual annotations, which also serves as a fine-tuning resource. 
Our experiments demonstrate the effectiveness of our pre-training strategy, the importance of scaling the pre-training dataset as well as the benefits of fine-tuning on smaller, high-quality data. Our approach not only sets the state of the art on the new manually annotated iGround dataset but also outperforms existing methods on the VidSTG, ActivityNet-Entities, GroundingYoutube and YouCook-Interactions benchmarks. 
We believe this work provides a strong foundation for future research in grounded video captioning.

\paragraph{Acknowledgments.}
This work was supported by the Ministry of Education, Youth and Sports of the Czech Republic through the e-INFRA CZ (ID:90140). This research also received the support of projects EXA4MIND, ELLIOT, CLARA and ERC FRONTIER, funded by the European Union’s Horizon Europe Research and Innovation Programme, under Grant Agreements N° 101092944, N° 101214398, N° 101136607  and N° 101097822. Views and opinions expressed are however those of the author(s) only and do not necessarily reflect those of the European Union or the European Research Council. Neither the European Union nor the granting authority can be held responsible for them. Furthermore, this work was funded in part by the French government under management of Agence Nationale de la Recherche as part of the “France 2030” program, reference ANR-23-IACL-0008 (PR[AI]RIE-PSAI projet), and the ANR project VideoPredict (ANR-21-FAI1-0002-01). Cordelia Schmid would like to acknowledge the support by the K\"orber European Science Prize.

{
    \small
    \bibliographystyle{ieeenat_fullname}
    \bibliography{main}
}

\clearpage

\appendix
\section*{Appendix}
\label{sec:appendix}

To complement the main paper, this appendix assembles additional results, analyses, and implementation details. Sections~\ref{sec:qualitative} and \ref{sec:quantitative} provide additional qualitative visualizations and expanded quantitative metrics, respectively. Section~\ref{sec:limitations} discusses the model and data limitations. Section~\ref{sec:dataset_stats} reports comprehensive dataset statistics. Section~\ref{sec:GROC_details} details the GROVE architecture and training setup, while Section~\ref{sec:automatic_annotation_method_details} describes our automatic annotation pipeline. Section~\ref{sec:annotation_protocol} outlines the human-annotation protocol, and Section~\ref{sec:prompts} lists the exact prompts used to curate spatio-temporally grounded captions.

\section{Additional qualitative results}
\label{sec:qualitative}
\begin{figure*}[htpb]
  \vspace*{-5mm}
  \includegraphics[width=\textwidth]{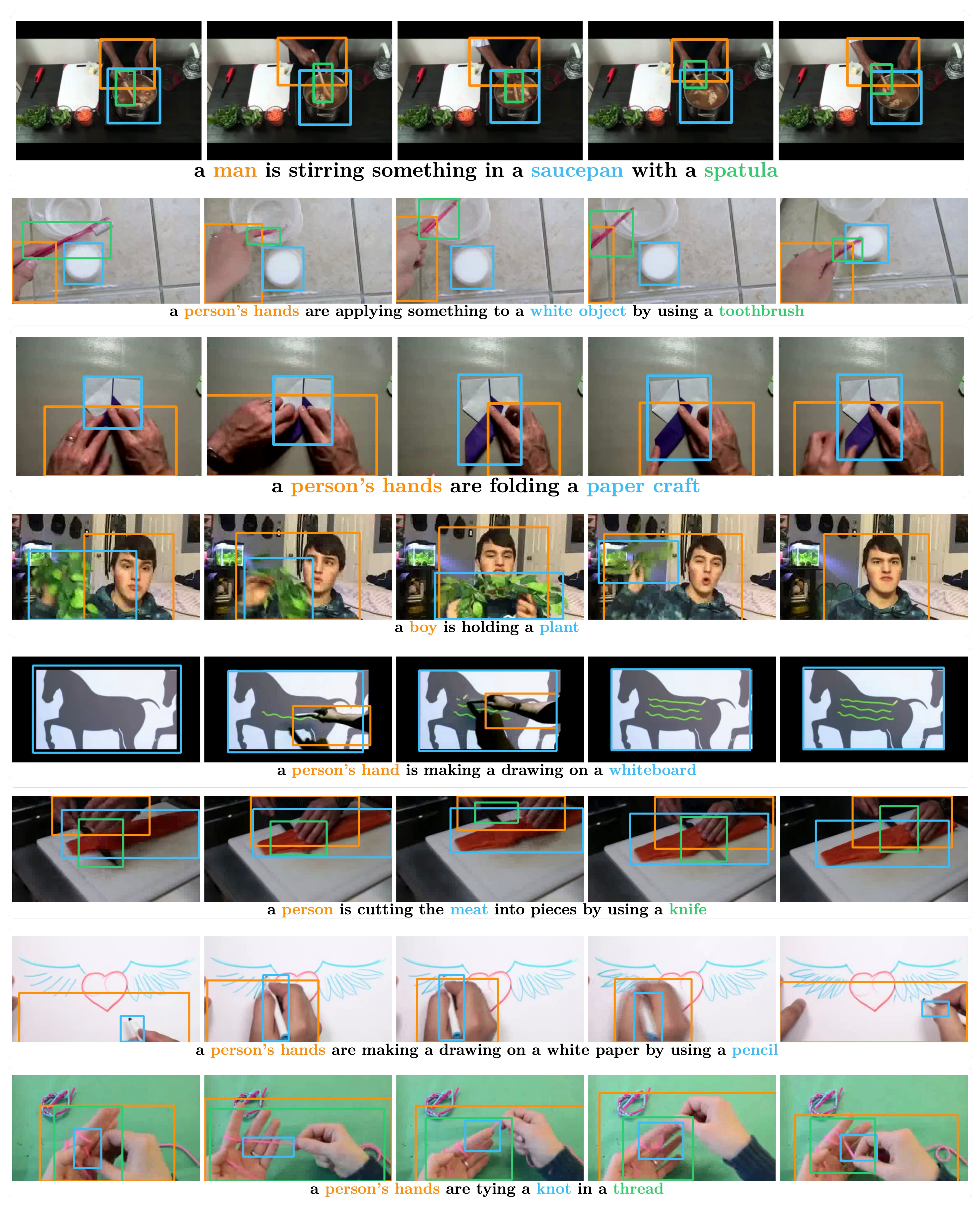}  
  \caption{Qualitative results of our GROVE model on the (unseen) iGround test set.
  The colour-coded sentence fragments are spatio-temporally localised in the video with the bounding boxes colour coded with the same colour. The results demonstrate that: (i) our model can localise even small objects such as a pen or a tooth brush; (ii) objects are consistently labelled across frames despite changes of viewpoint or scale; (iii) the model focuses on the human and the interacted objects; (iv) the model can successfully ground multiple objects in the video.
  {\bf Additional results are shown in the supplementary video} (available at \url{https://ekazakos.github.io/grounded_video_caption_generation/}).
  } 
  \label{fig:qualitative_supp1}
\end{figure*}

\begin{figure*}[htpb]
  \vspace*{-5mm}
  \includegraphics[width=\textwidth]{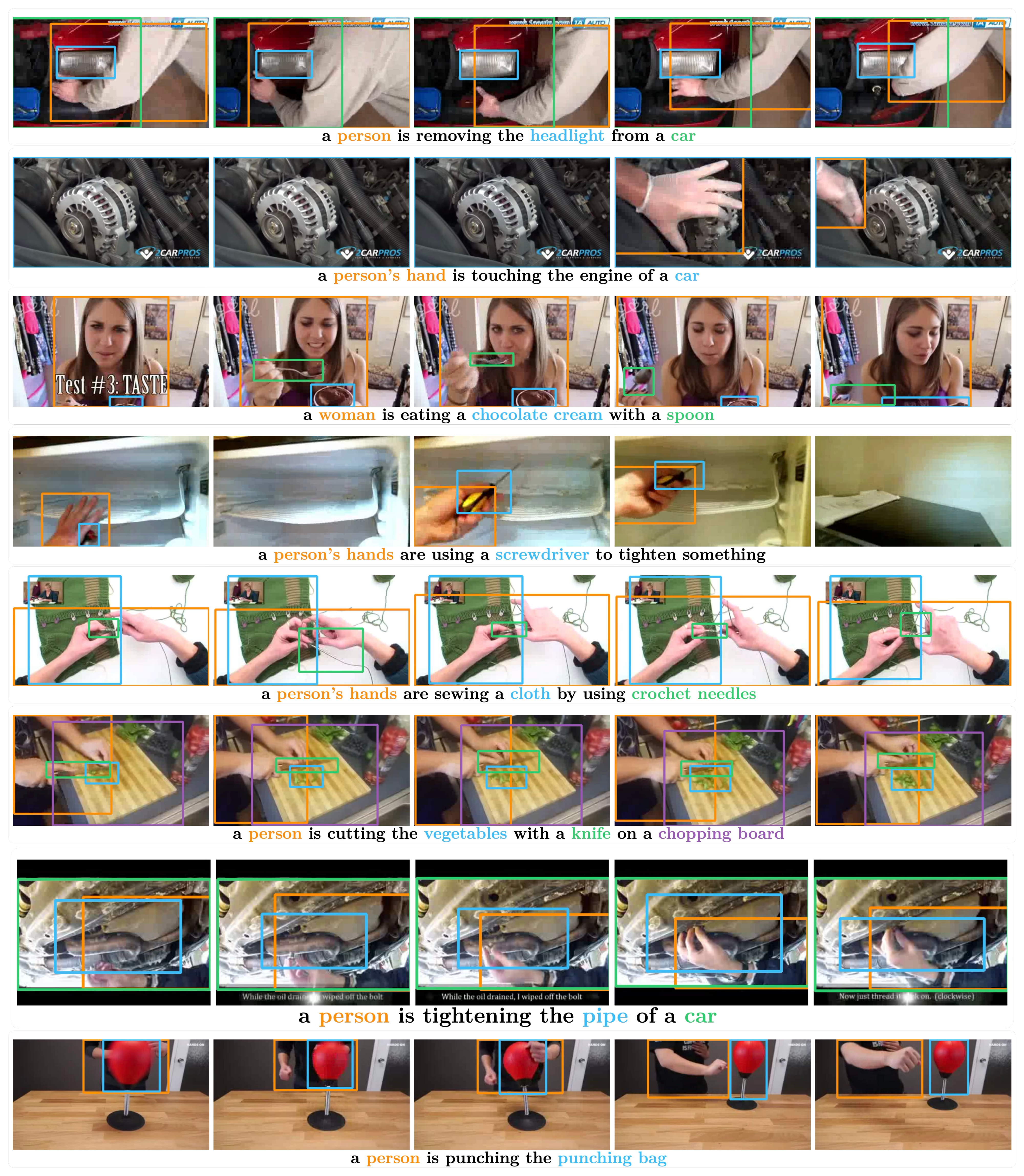}  
  \caption{Additional qualitative results of our GROVE model on the (unseen) iGround test set.
  The colour-coded sentence fragments are spatio-temporally localised in the video with the bounding boxes colour coded with the same colour. In addition to the model's properties discussed in Fig.~\ref{fig:qualitative_supp1}, GROVE is capable of predicting whether an object is present in a certain frame via the temporal objectness head; in the second example there are no bounding box predictions for the hand in the first three frames while in the fourth example there are no predictions for the hand and the screwdriver in the second and fifth frame. {\bf Additional results are shown in the supplementary video} (available at~\url{https://ekazakos.github.io/grounded_video_caption_generation/}). 
  } 
  \label{fig:qualitative_supp2}
\end{figure*}

\begin{figure*}[htpb]
  \vspace*{-5mm}
  \includegraphics[width=\textwidth]{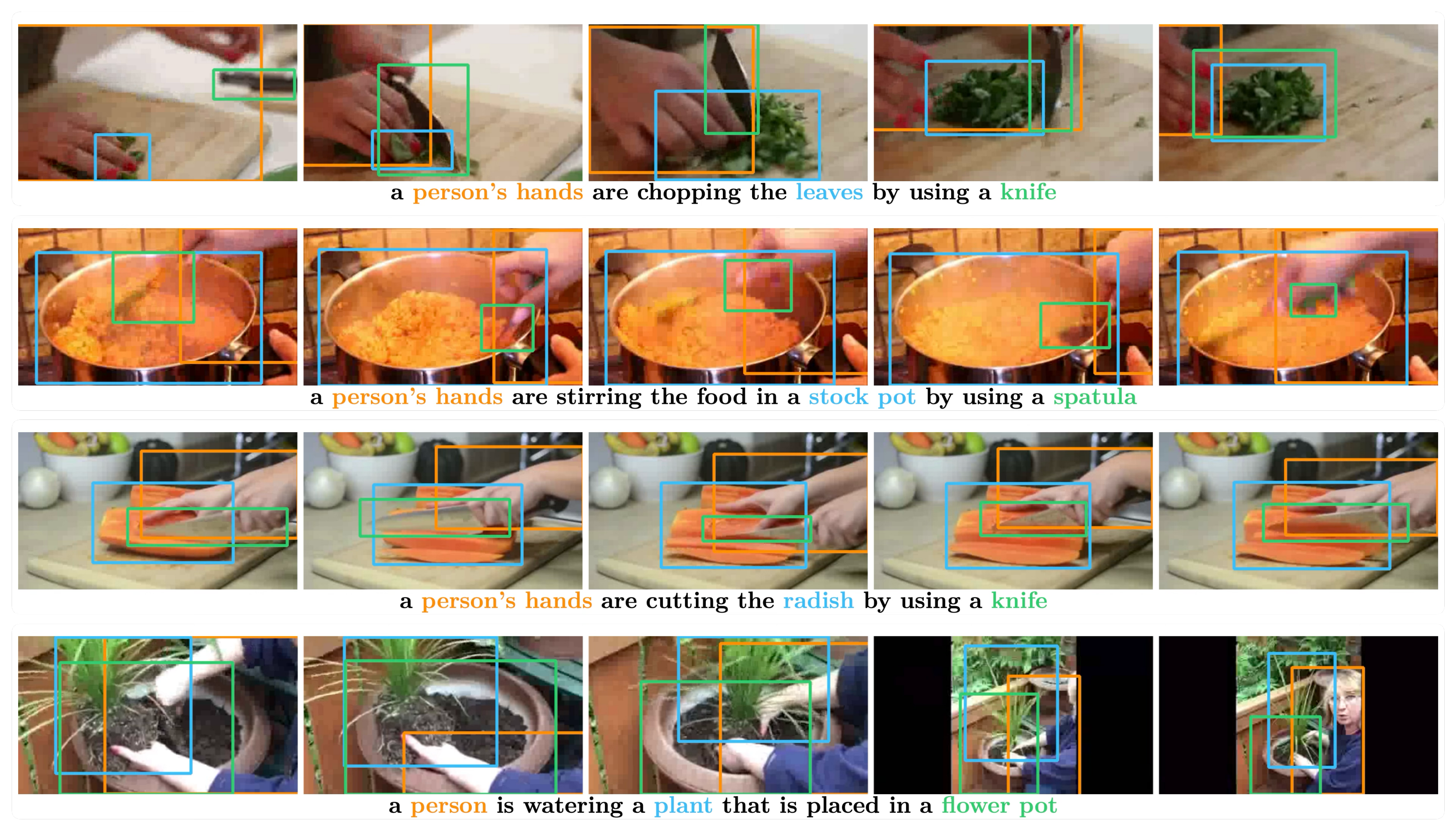}  
  \caption{\chg{{\bf Qualitative results for the main failure modes of our GROVE model on the (unseen) iGround test set.}
  The colour-coded sentence fragments are spatio-temporally localised in the video with the bounding boxes colour coded with the same colour. We identify four main failure modes: (i) temporal objectness mispredicts the presence of an object (first row, last frame for the knife), (ii) inaccurate predictions of object location (second row, third and last frames for the spatula), (iii) misclassification of object (third row, model predicts ``radish'' for the pumpkin), and (iv) misclassification of action (last example, model predicts ``watering'' for planting).}
  } 
  \label{fig:qualitative_fail}
\end{figure*}

Figures~\ref{fig:qualitative_supp1}~and~\ref{fig:qualitative_supp2} show qualitative results of our GROVE model (Section~\ref{sec:groc_model}), pre-trained on the HowToGround1M dataset and finetuned on the iGround training set (2013 examples). The results are shown on the iGround test set. In the figures' captions we discuss some of the benefits of our model.
Additional qualitative results showcasing the predictions of our approach overlaid over the input videos are shown in the {\bf supplementary video} (available at \url{https://ekazakos.github.io/grounded_video_caption_generation/}). \chg{Figure~\ref{fig:qualitative_fail} shows the main failure modes of our model.}

\section{Additional quantitative results}
\label{sec:quantitative}

\chg{\textbf{Detailed analysis for the ablations of automatic annotation}.}
\chg{We replace each stage of our automatic annotation method with an alternative. Results are shown in Table~\ref{tab:results_alternative_stage1}. In Stage 1, we replace the still-image model~\cite{hanoona2023GLaMM} with an alternative still-image grounded caption generation method. This approach leverages GIT~\citep{wang2022git} for frame-level captioning, Llama3~\citep{dubey2024llama3herdmodels} for extracting noun phrases from the caption, and OWLv2~\citep{minderer2024scalingopenvocabularyobjectdetection} for their bounding box localisation within each frame. We call this alternative ``b. Alt. Stage 1 (F)''. We also evaluate a video-level variant ``c. Alt. Stage 1 (V)'', where we replace the GIT captioner with VideoLlama3~\cite{zhang2025videollama3}. To ablate Stage 2 (``d. Alt. Stage 2''), we provide the LLM with full captions from Stage 1 instead of extracting Subject-Verb-Object triplets from the caption to assess the impact of additional context. To ablate Stage 3 (``e. Alt. Stage 3''), we incorporate CoTracker3~\cite{karaev24cotracker3}, a SOTA visual point tracking method to provide temporal association of bounding boxes across frames. Using 5 uniformly sampled frames and their bounding box predictions from Stage 1, we track objects in between with CoTracker3 and associate the resulting tracks with noun phrases from the caption.}

\chg{Results are reported in Table~\ref{tab:results_alternative_stage1}, where we compare the alternative automatic annotation methods on the iGround validation set. The frame-level alternative Stage 1 (row b.) performs better in captioning due to GIT's superior performance but performs noticeably worse for grounding. This is because our Stage 1 still-image grounding model~\cite{hanoona2023GLaMM} is explicitly trained for grounding, unlike GIT, Llama3, and OWLv2, which are not trained jointly and may underperform due to various factors--such as Llama3 extracting non-groundable noun phrases or OWLv2 missing objects. 
A similar trend is observed for the video-level alternative Stage 1 (row c.).
Compared to our proposed method, the alternative Stage 2 (row d.) underperforms across all metrics except AP50. This is because the full-caption input yields fewer
predictions as the LLM trims its output to the most salient
objects, reducing recall but improving precision–hence the
slightly higher AP50. In contrast, the SVO-based input in our proposed automatic annotation method leads to slightly longer captions (12 vs.11 words) with more noun phrases (3.3 vs. 3.0), leading to more object predictions and higher recall. 
This reflects a typical precision-recall trade-off. The alternative Stage 3 (row e.) underperforms in grounding due to tracker drift caused by abrupt viewpoint changes. Overall, on average (the last column in Table~\ref{tab:results_alternative_stage1}), our proposed method achieves the best performance.}

\noindent\chg{\textbf{Comparison with the state of the art on YouCook-Interactions and GroundgYouTube datasets}. In Table~\ref{tab:other_datasets}, we evaluate GROVE on YouCookInteractions and GroundingYouTube datasets, outperforming the previous SOTA by large margins.}

\begin{table}[h]
\centering
\resizebox{\columnwidth}{!}{%
\begin{tabular}{lcc}
\toprule
\textbf{Method} & \textbf{YouCook-Interactions} & \textbf{GroundingYouTube} \\
\midrule
What When and Where (S3D)~\cite{Chen_2024_CVPR} & 53.98 & 60.62 \\
What When and Where (CLIP)~\cite{Chen_2024_CVPR} & 58.35 & 56.98 \\
GROVE & \textbf{68.67} & \textbf{72.14} \\
\bottomrule
\end{tabular}
}
\caption{\chg{Comparison with SOTA on YouCook-Interactions~\cite{tanCOMMA2021} and GroundgYouTube\cite{Chen_2024_CVPR} datasets.}}
\label{tab:other_datasets}
\end{table}

\begin{table}[t]
\centering
\begin{smalltabularx}{Xlllll}
\toprule
& Method & METEOR & CIDER & AP50 & Recall\\
\midrule
\multirow{2}{*}{\rotatebox{90}{All}} & Auto. annotation & 12.3 & 31.7 & 26.9 & 19.3\\
 & GROVE & \textbf{19.7} & \textbf{92.6} & \textbf{42.0} & \textbf{26.9} \\
\midrule\midrule
\multirow{2}{*}{\rotatebox{90}{Hard}} & Auto. annotation & 08.1 & 07.3 & 22.3 & 14.7\\
 & GROVE & \textbf{14.4} & \textbf{41.3} & \textbf{36.0} & \textbf{18.9}\\
\bottomrule
\end{smalltabularx}
\caption{Results of our automatic annotation method (Auto. annotation) and the complete proposed model (GROVE) on the entire iGround validation set (All) and for a subset (about 10\% of data) with challenging similar referring expressions (Hard).}
\label{tab:ref_exp}
\end{table}

\section{Limitations}
\label{sec:limitations}

\chg{Although our proposed datasets and model advance the state of the art in grounded video captioning and spatio-temporal sentence grounding, they also reveal avenues for future exploration, stemming from the following limitations.}

\noindent\textbf{\chg{Scaling to long videos}}. \chg{Despite achieving state-of-the-art results on VidSTG by running inference in a sliding-window manner over videos up to three minutes, the training phase remains memory-bound: we can supply the model with only eight frames per clip. This is sufficient for the short clips in HowToGround1M and iGround (8-10 seconds), where eight frames corresponds to about 1 fps sampling. For VidSTG's much longer videos, however, uniform sampling of only eight frames introduces large gaps between frames and prevents the model from seeing fine-grained temporal dependencies during training. Closing this discrepancy will require methods that can train directly on larger frame spans or more efficient representations of extended temporal context (\textit{e.g.} memory).}

\noindent\textbf{\chg{Complex referring expressions}}. \chg{We  examined the iGround validation set and discovered that roughly $10\%$ of its videos contain more than one object whose referring expressions (and appearance) are highly similar. We designate this challenging portion of the data as the ``Hard'' subset. In Table~\ref{tab:ref_exp}, we compare GROVE with our automatic annotation method in this subset. Although GROVE still surpasses the automatic annotation method on this subset, the marked drop in performance of both methods on the ``Hard'' subset (comparing to ``All'', \textit{i.e.} the full validation set) reveals that reliably disambiguating closely related referring expressions remains still a challenge. These results suggest opportunities to refine both the model architecture and the automatic annotation method to better handle such fine-grained cases.}

\section{Dataset Statistics}
\label{sec:dataset_stats}

\begin{figure}[ht]
  \centering
  \begin{subfigure}[b]{\linewidth}
    \includegraphics[width=\linewidth]{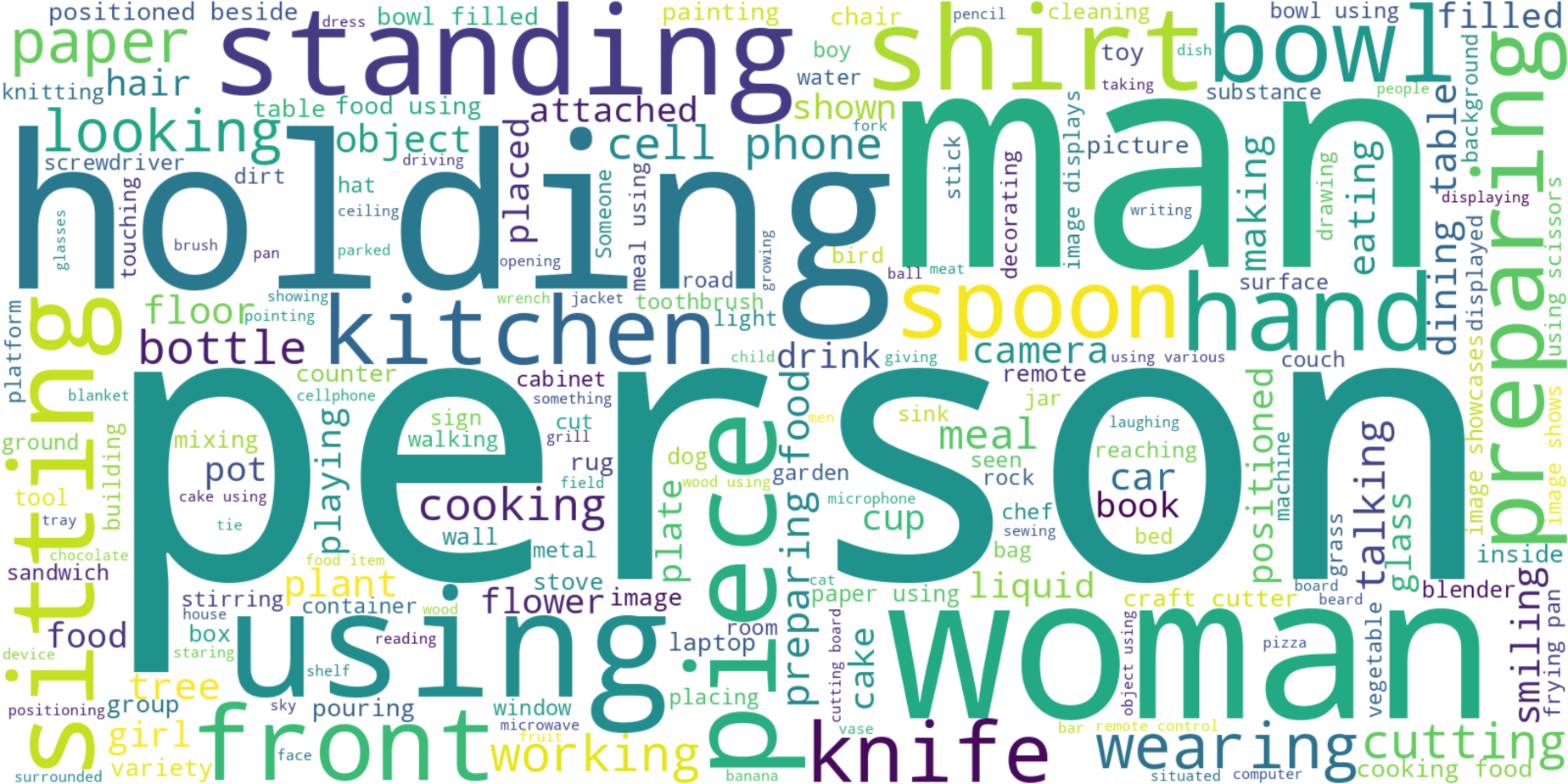} 
    \caption{}
    \label{fig:subfig1}
  \end{subfigure}
  \vspace{0.1cm} 
  \begin{subfigure}[b]{\linewidth}
    \includegraphics[width=\linewidth]{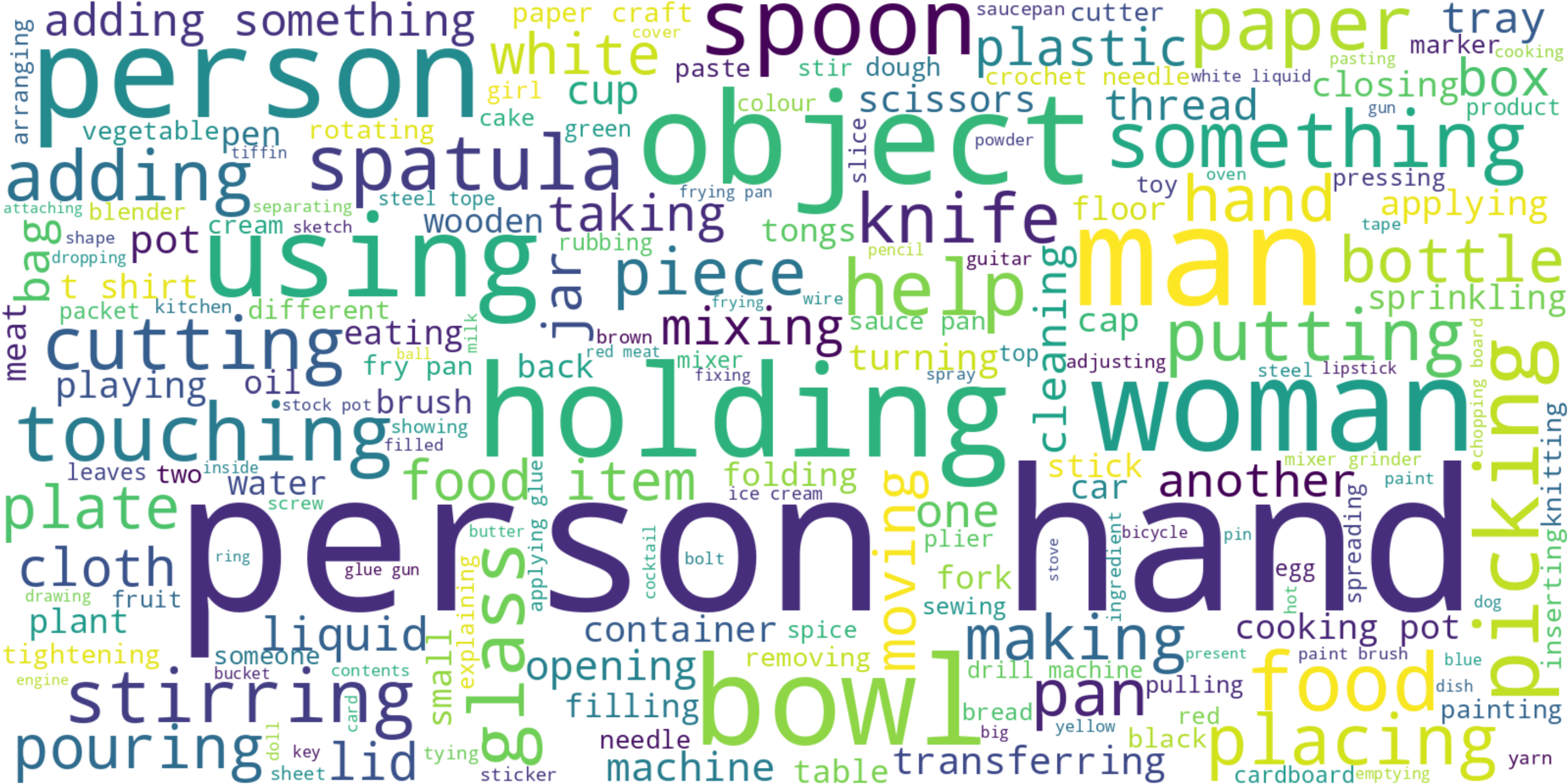} 
    \caption{}
    \label{fig:subfig2}
  \end{subfigure}
  \caption{Word cloud for (a) HowToGround1M dataset and (b) iGround dataset.}
  \label{fig:wordcloud}
\end{figure}

Table~\ref{tab:dataset_stats} reports the statistics of both the HowToGround1M pre-training dataset and the iGround manually annotated set. Word clouds of the natural language descriptions from those datasets are shown in Figure~\ref{fig:wordcloud}.

\begin{table}[h]
\centering
\resizebox{\columnwidth}{!}{
\begin{tabular}{lcc}
\toprule
Statistic & HowToGround1M & iGround \\
\midrule
Avg num frames per video & 44.6 & 40.1 \\
Avg duration (seconds) & 7.9 & 8.0 \\
Avg num instances per video & 80.1 & 118.1 \\
Total num instances & 80,092,775 & 421,588 \\
Avg box width $\times$ height & 243.7 $\times$ 172.6 & 174.9 $\times$ 135.5 \\
Avg tube length (frames) & 6.4 & 29.0 \\
Avg caption length (words) & 12.1 & 15.4 \\
\bottomrule
\end{tabular}
}
\caption{Statistics of HowToGround1M and iGround datasets.}
\label{tab:dataset_stats}
\end{table}

\section{Details of the GROVE model}
\label{sec:GROC_details}

\noindent\textbf{Model architecture.} Figure~\ref{fig:groc_model} 
shows the different components of our approach. The Global Video Encoder, $\mathcal{V}_e(\cdot)$, outputs video features, $o_e$, which are pooled spatio-temporally, resulting in the video prompts. These are projected to a language embedding space with $VL(\cdot)$. The LLM, $\mathcal{LM}(\cdot)$, ingests a multimodal prompt consisting of video and language tokens. The LLM is prompted to generate a caption for the video by tagging the noun phrases that correspond to objects and appending them with detection tokens (shown with red and green in the LLM's generated caption in Figure~\ref{fig:groc_model}). 
The LLM's output hidden states that correspond to the generated caption are projected to queries (using $LQ(\cdot)$). The queries corresponding to the detection tokens are fed to the bounding box decoder $D(\cdot)$. The Grounding Video Encoder, $\mathcal{V}_g(\cdot)$, outputs fine-grained video features, which are also fed to the decoder. The decoder performs cross-attention frame-wise between the queries and the outputs of $\mathcal{V}_g(\cdot)$, $o_g$, which are used as keys/values. Finally, the prediction heads output bounding box predictions and temporal objectness scores for each object at each frame. This objectness score is used to predict the presence/absence of the object in each video frame and is of major importance for the grounded video caption generation task. Details about the visual backbones $\mathcal{V}_e(\cdot)$ and $\mathcal{V}_g(\cdot)$ as well as details about the LLM $\mathcal{LM}(\cdot)$ including the format of its multimodal inputs and its vocabulary are given next. 

\noindent\textbf{Projection layers}. We project the outputs of the Global Video Encoder and the output hidden states of the LLM with MLPs, $o_{p'} = VL(o_p)$ and $o_q = LQ(o_l)$, where $VL(\cdot)$ projects the visual features to an embedded language space, while $LQ(\cdot)$ projects the LLM's hidden states to queries. $o_{p'}$ is the LLM's visual input while $o_q$ is input to the bounding box decoder that is described next.

\noindent\textbf{Backbones}. GROVE consists of two video encoders and a multimodal LLM as its main backbones. 
The Global Video Encoder $\mathcal{V}_e(\cdot)$, takes as input a video $v \in \mathbb{R}^{T \times H1 \times W1}$ and produces an output $o_e \in \mathbb{R}^{T \times \frac{H1}{p} \times \frac{W1}{p}}$, where $p$ is the patch size of the underlying visual transformer. Its purpose is to provide a holistic representation of the video that will be ingested by the LLM. The Grounding Video Encoder $\mathcal{V}_g(\cdot)$, takes as input a video $v \in \mathbb{R}^{T \times H2 \times W2}$, where $W2>W1$ and $H2>H1$. It produces $o_g \in \mathbb{R}^{T \times \frac{H2}{p} \times \frac{W2}{p}}$. $o_g$ is used to ground phrases from the caption to the visual content, which is performed by the bounding box decoder that is described later. The input video to the Grounding Video Encoder is of larger spatial resolution than that of the Global Video Encoder for enhanced localisation capability. Finally, the LLM $\mathcal{LM}(\cdot)$ takes as input a multimodal sequence $s \in \mathbb{R}^{L \times D}$ and produces an output $o_l$ of the same size. Its input is of the form \texttt{The <video> provides an overview of the video. Could you please give me a description of the video? Please respond with interleaved bounding boxes for the corresponding parts of the answer.} \texttt{<video>} is replaced by the output of $\mathcal{V}_e(\cdot)$, and therefore the LLM ingests mixed language and visual tokens. We also augment the LLM's vocabulary with a detection token \texttt{<DET>}, prompting the model to generate responses with \texttt{<DET>} tokens by the phrases that correspond to objects to be detected in the video.

\noindent\textbf{Loss function}. Our loss function is a combination of a language modelling loss and losses relevant to video object detection. The language modelling loss is a Cross-Entropy loss applied on $o_l$. For object detection, we follow DETR~\citep{detr} and use a gIoU loss~\citep{Rezatofighi_2019_CVPR} and an L1 loss applied on $p_{bb}$. Different than \cite{detr}, the losses are applied per frame and summed over frames. Moreover, the losses are applied only to the objects that appear in the frame (rather than each object in the caption) using the ground-truth temporal objectness scores. The representation that we use for the bounding boxes is \texttt{[x,y,w,h]} and their coordinates are normalised with the dimensions of the video. Finally, we employ a binary cross-entropy loss on $p_{tobj}$. Our loss is, hence, defined as:
\begin{align}
    \mathcal{L}_{LM} &= CE(o_l)\\
    \mathcal{L}_{gIoU} &= gIoU(p_{bb}, gt_{bb})\\
    \mathcal{L}_{L1} &= L1(p_{bb}, gt_{bb})\\
    \mathcal{L}_{tobj} &= BCE(p_{tobj}, gt_{tobj})\\
    \mathcal{L} &= \lambda_{LM}\times \mathcal{L}_{LM}+\lambda_{gIoU}\times\mathcal{L}_{gIoU}\\
    &+\lambda_{L1}\times \mathcal{L}_{L1}+\lambda_{tobj}\times \mathcal{L}_{tobj}\text{,} 
\end{align}
where $gt_{bb}$ are the ground truth boxes and $gt_{tobj}$ are the ground truth objectness scores and $\lambda$ are the weights for the losses.

\noindent\textbf{Training/inference}. We realise the Global Video Encoder  $\mathcal{V}_e(\cdot)$ with a CLIP-L~\citep{clip} model with an input of 336$\times$336 and a patch size of 14.  The Grounding Video Encoder $\mathcal{V}_g(\cdot)$ is instantiated with a SAM~\citep{kirillov2023segment} encoder and the bounding box decoder $D(\cdot)$ is a SAM-based decoder, the same as in GLaMM~\citep{hanoona2023GLaMM}. 
The LLM $\mathcal{LM}(\cdot)$ is a Vicuna-7B model~\citep{vicuna2023}. During training we keep $\mathcal{V}_e(\cdot)$, $\mathcal{V}_g(\cdot)$ and $\mathcal{LM}(\cdot)$ frozen. $\mathcal{V}_g(\cdot)$ originally takes as input 1024$\times$ 1024 images. As this is too large to fit in memory for videos, we instead use 512$\times$512 video spatial resolution, while we interpolate the positional encodings of $\mathcal{V}_g(\cdot)$ and fine-tune them. Adapters are  3D spatiotemporal convolutional layers with a kernel of size $3\times3\times3$ and a stride of 1. We apply adapters to every 3 layers of $\mathcal{V}_e(\cdot)$ and to all global attention layers of $\mathcal{V}_g(\cdot)$. The bounding box head $h_{bb}$ is an MLP with two FC layers and a ReLU activation function in between, while the temporal objectness head $h_{tobj}$ is a linear layer. Both prediction heads employ a sigmoid activation function. We apply a threshold of 0.5 to the temporal objectness scores. Both the adapters and the prediction heads are randomly initialised. We use $T=8$ frames for the videos during both training and testing. During training we perform random sparse sampling of frames by splitting the video in 8 segments and randomly drawing a frame from each segment while during testing we pick the centre frame of each segment. 

We train GROVE for 20 epochs using a batch size of 128. We use a learning rate of $5\times10^{-5}$ with warmup for the first 100 training steps and linearly decay the learning rate for the rest of training. We do not apply any weight decay or spatial data augmentation. We use $\lambda_{LM}=1\text{,}\lambda_{gIoU}=\lambda_{L1}=\lambda_{tobj}=2$.

\paragraph{Details of VidSTG and ActivityNet-Entities experiments.}
For VidSTG~\cite{vidstg} and ActivityNet-Entities~\cite{zhou2019grounded}, we do not use the temporal objectness head. That is because in VidSTG the spatio-temporal tubes are continuous within the segments' boundaries, while ActivityNet-Entities provides annotations for a single frame per object and in the rest of the frames the objects might still be present but without annotation, and thus should not be modelled as absent. As the task in VidSTG entails predicting the spatio-temporal bounding boxes \textit{given} a short description, we provide the short descriptions as input to our GROVE model during both training and inference in a teacher-forcing setup. \chg{For evaluating GROVE on VidSTG \textit{without} observing any VidSTG data during training (GROVE with FT: \xmark, Table~\ref{tab:vidstg_interrogative}), we pre-train GROVE on HowToGround1M. Each HowToGround1M caption is rewritten--by prompting Llama-3--into both of VidSTG's sentence styles, declarative and interrogative. Every transformed sentence is then paired with a single bounding box per frame, chosen as the box of the first subject or object it mentions. This supervision reshapes HowToGround1M's annotation distribution to mirror VidSTG's, allowing GROVE to achieve strong performance without relying on any VidSTG training data.}

\section{Details of the automatic annotation method}
\label{sec:automatic_annotation_method_details}
\textbf{Multiple people in the video}. Our automatic annotation method can handle multiple subjects in a video as long as one of the two following conditions are met: a) the subjects are described with a distinct language, \textit{e.g.} `man with green jumper' and `man with blue shirt', or b) the subjects are within a Subject-Verb-Object relationship even when described with the same terms, \textit{e.g.} (`person', `dances', `with', `person') which would produce `A person dances with another person'. If neither conditions are met, the caption aggregation (Stage 2) may merge the two subjects into one. 

\noindent\textbf{Association of verbs and objects} is naturally performed through the Subject-Verb-Object triplets. For example, given two relationships: (`man', `cuts', `onions') and (`woman', `stirs', `food', `in', `pot'). The LLM-based caption aggregation step (Stage 2) has sufficient information to associate the man with the action of cutting the onions and the woman with stirring the food. 

\noindent\textbf{Additional details of Stage 3}. We provide additional details of the procedure of Stage 3 using the example from Figure~\ref{fig:pseudolabelling}, right. The object in the woman's hands is described as `a green beverage' and `a glass of green liquid' across different frames. Stage 2 has provided the video-level noun phrases `a woman' and `a beverage'. Stage 3 is formulated as a classification problem where each one of `a green beverage' and `a glass of green liquid' are the inputs to be classified in one of the classes \{`a woman', `a beverage', $\emptyset$\} and thus associated with the right bounding box. The class $\emptyset$ represents the ``None'' class, \textit{i.e}. when an input does not belong to any of the known classes and it is useful for noisy inputs.

\begin{figure*}[h!]
\centering
\noindent\fbox{%
\parbox{\textwidth}{%
\textbf{Annotation Guidelines:}
\begin{enumerate}
    \item Video Selection:
    \begin{itemize}
        \item You will be provided with a larger set of videos than needed.
        \item Your first task is to select clips that are considered `interesting' based on criteria that will be discussed further. An `interesting' video typically includes dynamic events or actions that are clear and distinguishable despite the low video quality. In those events/actions people usually interact with objects, e.g. `A man is cutting an onion using a knife'. `Non-interesting' events are typically static, e.g. a person simply standing/sitting and talking. Non-interesting events are also events with ambiguous actions taking place, i.e. generic/abstract actions that cannot be described concisely or actions for which the annotator is unsure about what is happening in the video.
    \end{itemize}
    \item Video Annotation:
	\begin{itemize}
    \item For each selected video clip, write a concise, one-sentence description of the main event taking place in the clip. If the action is too complex, use at most two sentences for describing it, but prioritise one-sentence descriptions.
    \item Focus only on the objects that humans interact with rather than describing densely every object in the scene.
    \item To enrich the language descriptions, also describe properties of objects such as color, shape, etc, e.g. `blue cup' or `red onion'. It is not strictly necessary to always describe the object’s property but only when deemed important by the annotator.
    \item When you are unsure about the object being used, you can simply describe it as `object'. If object is unknown but the category of the object is known, please describe the object using its category, e.g. `food'.
    \item When there are two or more humans in the scene, use one of their characteristics to distinguish them, e.g. `the woman in the red shirt standing next to the woman in the green shirt is putting a strawberry on a cocktail glass'. 
    \item If there are multiple actions happening consecutively, describe all of them and their associated objects. E.g. `a person is doing action-1 using object-1, then doing action-2 with an object-2'. As shown in the example, you can use ‘then’ for connecting temporally adjacent actions.
    \item Provide bounding boxes for humans/different objects mentioned in your description. These bounding boxes should be applied to all frames where the objects are visible.
    \item Label each bounding box with a short phrase directly from your sentence description (e.g., `a brown dog', `person\'s hands').
    \item It is not necessary that each object appears in each frame of the video. For example, a person might be using a tool, then leaving it down and using another tool. In this case, you would annotate with bounding boxes the first tool for the first half of the video and the second tool for the second half. Another common case is that objects or the person might disappear and then reappear. In this case, again all instances of the object must be annotated, so you should be careful about objects leaving the scene as they might enter the scene again later. 
    \item If there are many small objects, e.g. mushrooms in a pan, use a single bounding box labelled as `mushrooms'.
    \item There are cases where two or more bounding boxes are needed for objects of the same type: a) one bounding box for each human hand when both are used to perform an action, b) one bounding box for each tool/container/appliance etc of the same type that the human is using, e.g. when they are placing food in two dishes, or pouring the content of a shaker in two cocktail glasses.
    \item Descriptions: Must be accurate and written in fluent English. Suitable for either native speakers or highly proficient English speakers.
    \item Bounding Boxes: Ensure that bounding boxes accurately encompass the objects for the entirety of their visibility within the clip. The bounding boxes should be consistent and smooth across frames, maintaining size and position as closely as possible given the movement of the object and video quality. An exception is when there are abrupt viewpoint changes of the camera, which might result in objects abruptly changing position and size across neighbouring frames.
\end{itemize}
\end{enumerate}
}}
\caption{Annotation guidelines for the manually annotated iGround dataset.}
\label{fig:annot_guidelines}
\end{figure*}

\section{Protocol for human annotations}
\label{sec:annotation_protocol}

In Figure~\ref{fig:annot_guidelines}, we describe the annotation guidelines for annotating the training/validation/test sets of the iGround dataset.

The annotation criteria have been extensively discussed with the annotation provider  and  the annotators have been trained based on those criteria prior to commencing the annotation process. We have also performed a pilot annotation project with the annotation provider on 10 video clips with several rounds of careful checking and feedback. Moreover, the annotation provider performed regular quality reviews on the annotations to ensure that the annotation criteria have been met.

\begin{figure*}[htbp]
\centering

\begin{tcolorbox}[colback=gray!10, colframe=gray!80, title=\textbf{System Instructions}, myboxstyle]
Generate a dynamic, video-level description based on frame-level inputs. The inputs include actions performed in individual frames in the form of Subject-Verb-Object (SVO) triplets along with prepositions and prepositional objects. The SVO triplets describe how actions are performed and how objects interact. Your output should be a concise narrative in 1 sentence, focusing on the most salient actions depicted across the frames. Enclose the exact text of relevant objects within \texttt{<p></p>} tags.

\textbf{Input format}:
\begin{verbatim}
[[`subject': `subject_text', `verb': `action_text', `object': `object_text', 
`prepositions_objects': [('preposition', `prepositional_object')],],]
\end{verbatim}

\textbf{Output format}:
\begin{verbatim}
A Python dictionary with a key `CAPTION', and as a value a dynamic description of the video content.
\end{verbatim}

Infer motion from static descriptions. E.g. `image shows a person holding a spoon and a bowl' implies `person is stirring food in a bowl'. Enclose the human and the most frequent object that is used to perform the action within \texttt{<p></p>} tags. If there is no human, enclose the two most frequent objects within \texttt{<p></p>} tags.
\end{tcolorbox}

\begin{tcolorbox}[pastelPink, title=\textbf{User Input 1}]
\textbf{SVO}: 
\begin{verbatim}
[[`image', `shows', `cup'], [`bowl', `is']], 
[[`person', `holding', `spoon'], [`spoon', `is', `bowl'], 
[[`image', `shows', `spoon', (`inside', `bowl')]], 
[[`person', `seen'], [`person', `holding', `spoon'], [`spoon', `used'], 
 [`spoon', `stir', `food', (`in', `bowl')]], 
[[`person', `holding', `spoon'], [`spoon', `is', `bowl']], 
[[`person', `holding', `spoon'], [`spoon', `is', `bowl']], 
[[`person', `holding', `spoon'], [`spoon', `is', `bowl']], 
[`image', `shows', `spoon', (`in', `bowl')]], 
[[`image', `shows', `bottle'], [`bottle', `positioned', (`beside', `bowl')]], 
[[`image', `shows', `bottle'], [`bottle', `positioned', (`beside', `cup')]], 
[[`image', `shows', `bottle'], [`image', `placed', (`on', `counter')], 
 [`bottle', `positioned', (`beside', `bowl')]]]
\end{verbatim}
\end{tcolorbox}

\begin{tcolorbox}[pastelGreen, title=\textbf{Assistant Response 1}]
\begin{verbatim}
{`CAPTION': `<p>A person</p> is stirring <p>food in a bowl</p> using a spoon'}
\end{verbatim}
\end{tcolorbox}

\begin{tcolorbox}[pastelPink, title=\textbf{User Input 2}]
\textbf{SVO}: 
\begin{verbatim}
[[`hand', `using', `cutting board']], 
[[`woman', `using', `cutting board'], [`woman', `make', `craft project']], 
[[`child', `using', `craft cutter'], [`child', `cut', `object']], 
[[`child', `using', `craft cutter'], [`child', `cut', `paper']], 
[[`woman', `using', `craft cutter'], [`woman', `cut', `object']], 
[[`woman', `using', `scissors pair'], [`woman', `cut', `piece', (`of', `paper')]], 
[[`hand', `using', `scissors pair'], [`hand', `cut', `piece', (`of', `paper')]], 
[[`woman', `using', `scissors pair'], [`woman', `cut', `piece', (`of', `paper')]], 
[[`woman', `using', `craft cutter'], [`woman', `cut', `object']], 
[[`woman', `using', `craft cutter'], [`woman', `cut', `plate']]]
\end{verbatim}
\end{tcolorbox}

\begin{tcolorbox}[pastelGreen, title=\textbf{Assistant Response 2}]
\begin{verbatim}
{`CAPTION': `<p>A woman</p> is cutting <p>an object</p> using a craft cutter'}
\end{verbatim}
\end{tcolorbox}

\begin{tcolorbox}[pastelPink, title=\textbf{New User Input}]
\textbf{SVO}: \verb|{input_svo}|
\end{tcolorbox}

\caption{The full prompt for Stage 2 (Video-level caption aggregation) of our automatic annotation approach  (Section~\ref{sec:pseudolabeling}).}
\label{fig:stage2}
\end{figure*}

\begin{figure*}[htbp]
\centering

\begin{tcolorbox}[colback=gray!10, colframe=gray!80, title=\textbf{System Instructions}, myboxstyle]
You are tasked with classifying humans and objects to a set of given categories. 

\textbf{Input format}: 
\begin{verbatim}
Human/Object (string), set of categories (lists of strings). 
\end{verbatim}

\textbf{Output format}: 

\begin{verbatim}
A Python dictionary with a key `CATEGORY', and as a value the predicted category of the human/object.
\end{verbatim}

Use `None' if the human/object doesn`t belong to any of the categories. DO NEVER classify a human as the object category and vice versa.   

\end{tcolorbox}

\begin{tcolorbox}[pastelPink, title=\textbf{User Input 1}]
\textbf{Input}: `person`\newline
\textbf{Categories}: [`a woman', `her hair']
\end{tcolorbox}

\begin{tcolorbox}[pastelGreen, title=\textbf{Assistant Response 1}]
\begin{verbatim}
{`CATEGORY': `a woman'}
\end{verbatim}
\end{tcolorbox}

\begin{tcolorbox}[pastelPink, title=\textbf{User Input 2}]
\textbf{Input}: `table'\newline
\textbf{Categories}: [`a person', `a bowl']
\end{tcolorbox}

\begin{tcolorbox}[pastelGreen, title=\textbf{Assistant Response 2}]
\begin{verbatim}
{`CATEGORY': `None'}
\end{verbatim}
\end{tcolorbox}

\begin{tcolorbox}[pastelPink, title=\textbf{User Input 3}]
\textbf{Input}: `a piece of food on a plate'\newline
\textbf{Categories}: [`a woman', `a meal']
\end{tcolorbox}

\begin{tcolorbox}[pastelGreen, title=\textbf{Assistant Response 3}]
\begin{verbatim}
{`CATEGORY': `a meal'}
\end{verbatim}
\end{tcolorbox}

\begin{tcolorbox}[pastelPink, title=\textbf{User Input 4}]
\textbf{Input}: `a hand'\newline
\textbf{Categories}: [`a person', `food on a plate']
\end{tcolorbox}

\begin{tcolorbox}[pastelGreen, title=\textbf{Assistant Response 4}]
\begin{verbatim}
{`CATEGORY': `a person'}
\end{verbatim}
\end{tcolorbox}

\begin{tcolorbox}[pastelPink, title=\textbf{User Input 5}]
\textbf{Input}: `a man in a white shirt and black apron is also present'\newline
\textbf{Categories}: [`a person', `food']
\end{tcolorbox}

\begin{tcolorbox}[pastelGreen, title=\textbf{Assistant Response 5}]
\begin{verbatim}
{`CATEGORY': `a person'}
\end{verbatim}
\end{tcolorbox}

\begin{tcolorbox}[pastelPink, title=\textbf{New User Input}]
\textbf{Input}: \verb|{input_object}|\newline
\textbf{Categories}: \verb|{input_categories}|
\end{tcolorbox}

\caption{The full prompt for Stage 3 (Temporally consistent bounding box annotation) of our automatic annotation approach (Section~\ref{sec:pseudolabeling}).}
\label{fig:stage3}
\end{figure*}

\section{Prompts for automatic
curation of spatio-temporally grounded captions}
\label{sec:prompts}

The full prompt for the {\bf Stage 2 (Video-level caption aggregation)} of our automatic annotation approach  (Section~\ref{sec:pseudolabeling}) is shown in Figure~\ref{fig:stage2} and the full prompt for {\bf Stage 3 (Temporally consistent bounding box annotation)} in Figure~\ref{fig:stage3}.

\end{document}